\documentclass[sigconf,nonacm]{acmart}

\AtBeginDocument{%
  }

\usepackage{amsmath}
\usepackage{booktabs}
\usepackage{enumitem}
\usepackage{multirow}
\usepackage{xspace}
\usepackage{tikz}
\usetikzlibrary{arrows.meta,positioning,shapes.geometric,fit,calc}
\usepackage{algorithm}
\usepackage{algpseudocode}

\newcommand{\apa}{APA\xspace}                 
\newcommand{\ppv}{PPV\xspace}                 
\newcommand{\bench}{ActiveWebPref\xspace}     
\newcommand{\ask}{\textsc{Ask}\xspace}
\newcommand{\defer}{\textsc{Defer}\xspace}
\newcommand{\skipa}{\textsc{Skip}\xspace}
\newcommand{\E}{\mathbb{E}}

\usepackage{float}
\usepackage{placeins}
\usepackage{fvextra}
\DefineVerbatimEnvironment{PromptText}{Verbatim}{fontsize=\scriptsize,breaklines=true,breakanywhere=true,breaksymbolleft={},formatcom=,baselinestretch=1.05}
\DefineVerbatimEnvironment{PromptJSON}{Verbatim}{fontsize=\scriptsize,breaklines=true,breakanywhere=true,breaksymbolleft={},formatcom=,baselinestretch=1.05}
\newtheorem{proposition}{Proposition}

\begin{document}

\title{Learning What to Ask for Tomorrow: Active Preference Acquisition for Long-Term Personalized Web Agents}

\author{Yiqi Wang$^{*}$, Zizhi Zeng, Jiaqi Zhang, Yiqun Duan, Mingkai Zheng, Taotao Cai}
\email{$^{*}$yiqi.wang.jennie@gmail.com}
\renewcommand{\shortauthors}{Wang et al.}

\begin{abstract}
Web agents increasingly serve the same user across repeated online tasks such as booking travel, shopping, and scheduling, so their long-term usefulness depends on how well they know the user's preferences.
Passive memory and personalization methods cannot obtain preferences the user never reveals, and clarification methods that value a question by its benefit to the current task miss preferences whose benefit on any single task is too small to justify asking but whose benefit over repeated reuse is not.
We study \emph{Active Preference Acquisition} (\apa): which missing preferences are worth the cost of asking, given their reuse in future tasks?
We define the \emph{Prospective Personalization Value} (\ppv) of a missing preference as its expected cumulative utility gain over future tasks minus the effort and disclosure-risk cost of asking, and characterize when cumulative reuse justifies acquisition despite insufficient single-task benefit.
\apa estimates \ppv from public task prototypes: the agent's own task solver measures each candidate's counterfactual gain, these gains are weighted by expected reuse, and questions are selected under asking cost and a budget. We evaluate it on \bench, a controlled longitudinal benchmark with hidden but queryable user profiles and exact task utilities.
In \bench, 34.4\% of budget-feasible valuable missing preferences are amortization-dependent at a horizon of 20 tasks; current-task value of information acquires 5.1\% of them and \apa 80.3\%.
Among the users holding such preferences, \apa's net utility exceeds that of current-task valuation by 0.263 at 20 tasks (Holm $p=0.021$), and under asking-cost caps matched to its own expenditure it achieves higher net utility than four heuristic selection baselines.
Preference acquisition should therefore be priced by reuse across future tasks; the remaining challenge is to acquire only the preferences whose reuse repays their cost.
\end{abstract}



\maketitle

\section{Introduction}
\label{sec:intro}

Web agents can now carry out realistic Web tasks such as online shopping and navigating real-world websites~\cite{yao2022webshop,deng2023mind2web,zhou2024webarena}, and they are moving from one-shot assistance to persistent service.
The same agent may book a user's conference travel in March, order office supplies in April, and reserve a team dinner in May.
Its usefulness over time then depends less on any single task than on how well it \emph{knows the user}: that the user avoids red-eye flights, keeps hotels under a per-night budget, and needs vegetarian options.
This paper studies that user-modeling problem for users of persistent Web agents: which missing preferences a persistent Web agent should acquire, and when doing so pays off.

Valuable preferences, however, need not appear in the interaction history.
Personalization methods condition on past behavior~\cite{salemi2024lamp} or infer profiles from what users have said~\cite{zhang2026personalagent}, and long-term memory systems decide how to extract, store, retrieve, and update observed information~\cite{packer2023memgpt,zhong2024memorybank,chhikara2025mem0,rasmussen2025zep,yang2026promem}.
These methods work with evidence that exists; a preference the user has never had reason to mention or to reveal through behavior cannot be extracted or retrieved.

Asking can fill such gaps, but asking is not free: each question costs the user effort, and each answer discloses personal information.
An agent therefore cannot simply ask about everything; in our experiments, asking about every unknown preference, or about whatever an LLM deems most worth knowing, yields lower net utility than not asking at all (\S\ref{sec:results}).

\begin{figure*}[t]
\centering
\includegraphics[width=.86\textwidth]{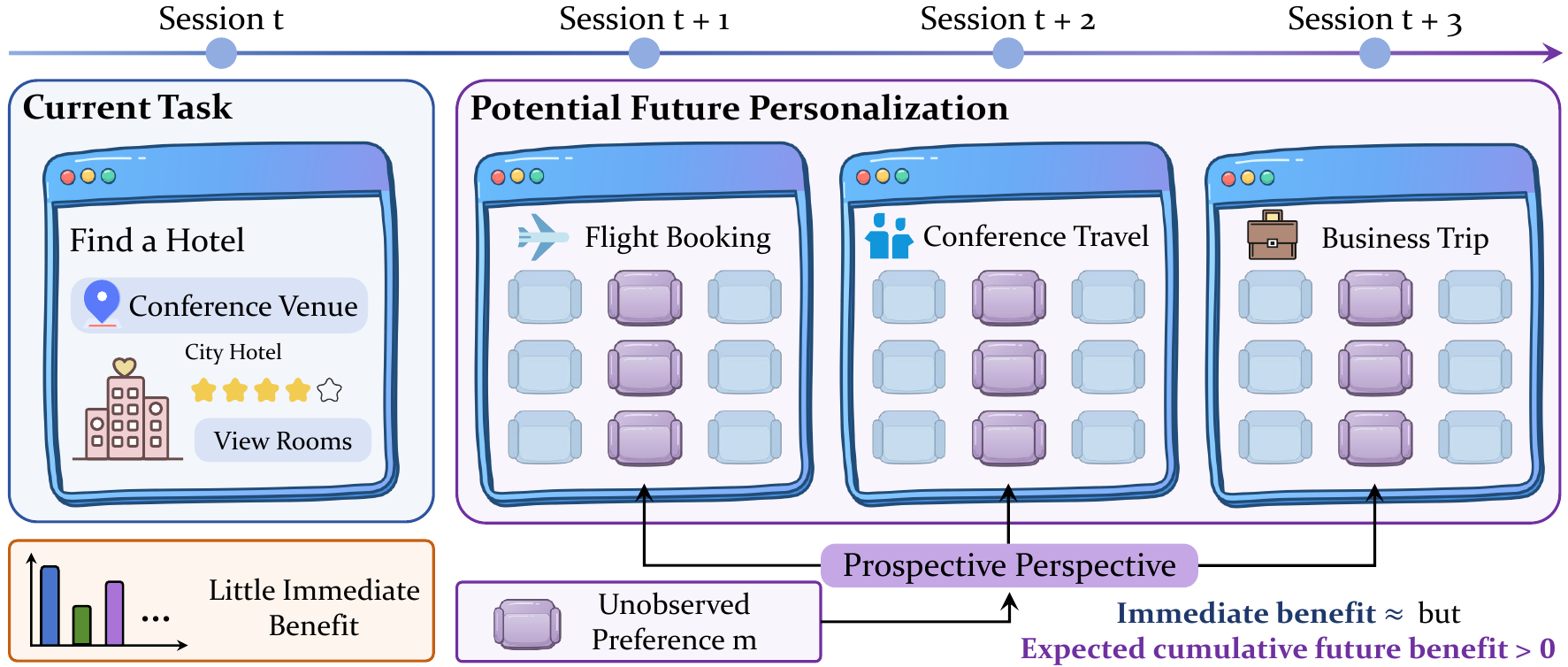}
\caption{A single use does not justify asking, but cumulative reuse does. In session $t$ the user asks for a hotel, and the seat preference $m$ is unknown and useless for this task. Each later seat selection gains from knowing $m$, but with illustrative numbers, a gain of 0.05 per use against an asking cost of 0.15, no single use justifies asking, so a criterion that values only the task at hand does not ask even when a flight is booked. Four uses gain 0.20, so acquiring $m$ once and reusing it pays off.}
\Description{Timeline from session t to t+3. In session t the user books a hotel; a missing seat preference has no immediate value. In sessions t+1 to t+3 flight, conference-travel and business-trip bookings each require selecting a seat; each use alone gains less than the cost of asking, but repeated reuse gains more.}
\label{fig:motivation}
\end{figure*}

Valuing each question by the current task does not resolve this either.
Current-task clarification and value-of-information (VoI) methods ask when a question improves the current task enough to justify its cost~\cite{rao2018learning,aliannejadi2019asking,zamani2020generating,dong2026voi}.
While booking a hotel, the agent can finish without the user's seat preference, so current-task VoI does not ask (Figure~\ref{fig:motivation}).
Why not wait and ask when a flight is booked?
Consider an illustrative preference that costs 0.15 to ask and improves each task that uses it by 0.05. At every single task, asking loses 0.10, so a criterion that values only the task at hand never asks; over four later uses the gain is 0.20, so acquiring it once has positive net value.
Such a preference is missed however often it is reused (Proposition~\ref{prop:amortization}).

An agent therefore needs to value acquisition across tasks:
\begin{quote}
\emph{Which currently unknown user preferences are worth acquiring, because their benefit across future tasks outweighs the cost of asking?}
\end{quote}
We call this problem \textbf{Active Preference Acquisition} (\apa).
At each interaction the agent holds a partial profile and can already complete the current task; it decides which missing preferences to ask about and stores the answers persistently.
Unlike cold-start elicitation or within-session clarification, \apa interleaves acquisition with ongoing service and values persistent attributes by their reuse across heterogeneous future tasks.
We define the \emph{Prospective Personalization Value} (\ppv) of a missing preference: its expected utility gain across future tasks, minus the effort and disclosure-risk cost of asking.
Two choices are involved, and we separate them: the \emph{valuation scope}, current benefit only or cumulative future benefit, and the \emph{timing} of a question, now or at the first relevant task. Our object of study is the valuation scope; asking early is a possible behaviour, not the claim. \ppv thus complements clarification rather than generalizing it.

We estimate \ppv by \emph{counterfactual reuse valuation}. For each candidate missing preference, the agent's own task solver computes how much knowing it would change utility on public task prototypes, and these gains are aggregated by the expected reuse of each task family; questions are then selected under asking cost, disclosure risk and a per-interaction budget.
To test when this helps and by how much, we build \bench, a controlled longitudinal testbed modeled on personalized Web services, with hidden but queryable user profiles, multi-domain task sequences, and exact task utilities. It isolates the value of preference acquisition, which existing memory benchmarks do not measure~\cite{wu2025longmemeval,maharana2024locomo}, and does not evaluate agents acting on live websites.

\medskip\noindent\textbf{Contributions.}
\begin{enumerate}[label=(\arabic*),leftmargin=*]
\item \textbf{Problem and diagnosis.} We formulate Active Preference Acquisition, define \ppv, and identify \emph{amortization-dependent} preferences, whose gain on every single task is below the asking cost while their cumulative gain exceeds it. Proposition~\ref{prop:amortization} shows that a criterion valuing only the current task never acquires a preference whose gain stays below the cost in every decision state it reaches; an initial-profile diagnostic gives a fixed, comparable set of such preferences for measuring this omission.
\item \textbf{Method.} \apa provides a prototype-based estimator of \ppv under the public task model: the agent's own task solver measures each candidate's counterfactual gain on public task prototypes, the gains are weighted by expected reuse, and an adaptive greedy policy selects questions under asking cost, disclosure risk and a per-interaction budget (\S\ref{sec:method}).
\item \textbf{Benchmark and findings.} \bench (\S\ref{sec:bench}) has 16 task families, hidden but queryable user profiles and exact task utilities; code, data and prompts are at \url{https://github.com/xiaoqi-7/Active-Preference-Acquisition}. In it, preferences that a single use does not justify but reuse does are common; current-task valuation rarely acquires them, and among the users holding them \apa achieves higher net utility at longer horizons, also on a separately generated diagnostic set enriched with them. Cost-matched comparisons support selective acquisition: \apa achieves higher net utility than heuristic selection under comparable asking expenditure (\S\ref{sec:results}).
\end{enumerate}

\section{Related Work}
\label{sec:related}

\paragraph{User modeling and personalization.}
Personalized systems condition generation or ranking on user profiles, given as explicit persona statements~\cite{zhang2018personalizing} or drawn from behavioral history~\cite{salemi2024lamp}.
Recent LLM agents continuously infer and refine a unified user profile from dialogue turns~\cite{zhang2026personalagent}, or update preference memories as user behavior evolves by fusing short- and long-term averages~\cite{sun2026pamu}.
These methods \emph{infer} preferences from evidence that already exists.
They cannot recover a preference that has never been expressed or revealed through behavior, and they do not decide which unobserved preferences are worth eliciting for future tasks.

\paragraph{Preference elicitation.}
Decision-theoretic preference elicitation frames question asking as a POMDP that trades question cost against the quality of a final decision~\cite{boutilier2002pomdp}.
Recommender systems ask new users to rate items in a cold-start interview~\cite{rashid2002getting}, possibly adapting the next item to earlier answers with a decision tree~\cite{golbandi2011adaptive}, and use active learning in collaborative filtering more broadly~\cite{elahi2016survey}.
Conversational recommenders decide within a session which item or attribute to ask about next~\cite{christakopoulou2016towards,sun2018conversational,lei2020estimation,lei2020cpr,deng2021unicorn,jannach2021survey}.
LLMs have also been used to elicit preferences through open-ended questions: GATE prompts models to elicit a user's intended task behavior~\cite{li2023eliciting}, and STaR-GATE fine-tunes a questioner to ask useful clarifying questions~\cite{andukuri2024stargate}.
\apa differs from this work in three ways.
(i)~\emph{When and why to ask.} Conversational elicitation asks in order to finish the \emph{current} session. Cold-start interviews do ask for future benefit, but in a single onboarding step before any task is served. In \apa, questions are interleaved with ongoing tasks that can already be completed, and each question is justified by its expected effect on future tasks.
(ii)~\emph{Scope.} Elicitation targets item ratings or attributes within a single catalog. \apa acquires persistent user attributes that are reused across heterogeneous task families, such as flights, hotels, dining, and scheduling.
(iii)~\emph{Valuation target.} Like prior elicitation, \apa weighs benefit against the cost of asking; the difference is that the benefit is counted across the heterogeneous future tasks that reuse the answer, rather than within one decision or session.
\apa builds on established value-of-information and preference-elicitation principles~\cite{howard1966information,boutilier2002pomdp}, operationalizing them for persistent preferences reused across heterogeneous tasks, and adds a benchmark in which the counterfactual value of an unasked preference can be measured.

\paragraph{Long-term memory for agents.}
Surveys organize agent memory by its sources, forms, and operations (writing, management, and reading)~\cite{zhang2025memorysurvey}.
Memory architectures give LLM agents persistent state through hierarchical context management~\cite{packer2023memgpt}, memory streams with reflection~\cite{park2023generative}, forgetting-aware stores~\cite{zhong2024memorybank}, scalable extraction and retrieval~\cite{chhikara2025mem0}, and temporal knowledge graphs~\cite{rasmussen2025zep}.
ProMem improves memory extraction itself: it extracts details, events, and relations separately, checks completeness to recover missed events, and verifies atomic facts~\cite{yang2026promem}.
Benchmarks such as LongMemEval~\cite{wu2025longmemeval} and LoCoMo~\cite{maharana2024locomo} test whether facts \emph{present in the history} can be recalled.
PrefEval tests whether models follow stated or implicit preferences~\cite{zhao2025prefeval}, and PersonaMem whether they track preferences that evolve across sessions~\cite{jiang2025personamem}; both find that preferences already expressed are often not followed.
These systems answer \emph{how to retain and use what is known}.
\apa addresses \emph{which unknown information should be learned}, and can run on top of any of them.

\paragraph{Clarification and value of information.}
Asking clarifying questions has been studied in information retrieval~\cite{aliannejadi2019asking,zamani2020generating} and in NLP, for example by ranking questions by expected value of perfect information~\cite{rao2018learning}.
For LLM agents, the IN3 benchmark annotates which details of a vague task are missing, and Mistral-Interact is trained to ask about them before executing~\cite{qian2024tellmemore}, and Ask-before-Plan studies clarification before travel planning~\cite{zhang2024askbeforeplan}.
Proactive Agent instead anticipates and offers help that the user has not requested~\cite{lu2025proactive}.
Dong et al.~\cite{dong2026voi} formalize when an agent should ask rather than act by weighing the expected utility gain of a question on the current decision against the cost it imposes on the user.
Two lines of work look beyond the current response: one labels response preferences by simulating future conversation turns, teaching models when to ask clarifying questions~\cite{zhang2025modeling}, and CollabLLM trains models with a multiturn-aware reward estimated from simulated future conversations~\cite{wu2025collabllm}.
Their horizon, however, is later turns of the \emph{same} conversation, whereas the horizon of \apa is future tasks across sessions.
Abstracting from these methods, current-task clarification scores a question about $m$ as
\[
V_{\text{cur}}(m)=\Delta U(\tau_t;m)-C(m),
\]
whereas \apa scores the same preference by its value over the tasks that follow,
\[
\ppv_T(m)=\E\Big[\textstyle\sum_{k=t+1}^{T}\Delta U(\tau_k;m)\Big]-C(m).
\]
The two scores answer different questions and are complementary.
Howard's formulation values information for a single decision problem~\cite{howard1966information}, and the non-myopic VoI of Krause and Guestrin~\cite{krause2005near} selects a budget-constrained set of observations jointly, rather than greedily, to reduce uncertainty in one fixed model.
These principles can accommodate long-term benefit; \apa instantiates them as a \emph{longitudinal personalization} decision.
The information is stored, and it pays off across many future tasks drawn from an uncertain task distribution.

\paragraph{Web agents.}
Benchmarks for Web agents such as WebShop~\cite{yao2022webshop}, Mind2Web~\cite{deng2023mind2web}, and WebArena~\cite{zhou2024webarena} pose each episode as a single, self-contained instruction, with no user state carried across tasks.
Personalized Web agents have recently been studied through PersonalWAB, which pairs simulated users' profiles and behavior histories with personalized search, recommendation, and review tasks and whose multi-turn track lets the agent clarify a request with a simulated user~\cite{cai2025personalwab}, and Persona2Web, which requires agents on the open Web to resolve ambiguous queries from long user histories~\cite{kim2026persona2web}.
Both personalize mainly from histories, and any questions serve the current request. Neither studies which missing preferences an agent should acquire for future tasks.
\section{Problem Formulation}
\label{sec:problem}

\paragraph{Persistent user--agent interaction.}
A user $u$ interacts with an agent over a sequence of tasks $\tau_1,\dots,\tau_T$.
Each task belongs to a \emph{task family} $f(\tau)\in\mathcal{F}$, such as flight booking, hotel selection, restaurant reservation, or meeting scheduling.
At step $t$ the agent observes the interaction history $H_t$ and maintains a persistent profile $P_t$, a set of attribute--value pairs such as $(\textit{seat},\textit{aisle})$.
The user has a hidden ground-truth profile $P^*_u$, and the agent's profile is a subset of it, $P_t\subseteq P^*_u$.
Given a task and a profile, the agent's task solver selects an action whose utility $U(\tau\mid P)\in[0,1]$ is determined by $P^*_u$.
In \bench, $U$ is computed exactly from the hidden profile (\S\ref{sec:bench}).

\begin{figure*}[t]
\centering
\includegraphics[width=\textwidth]{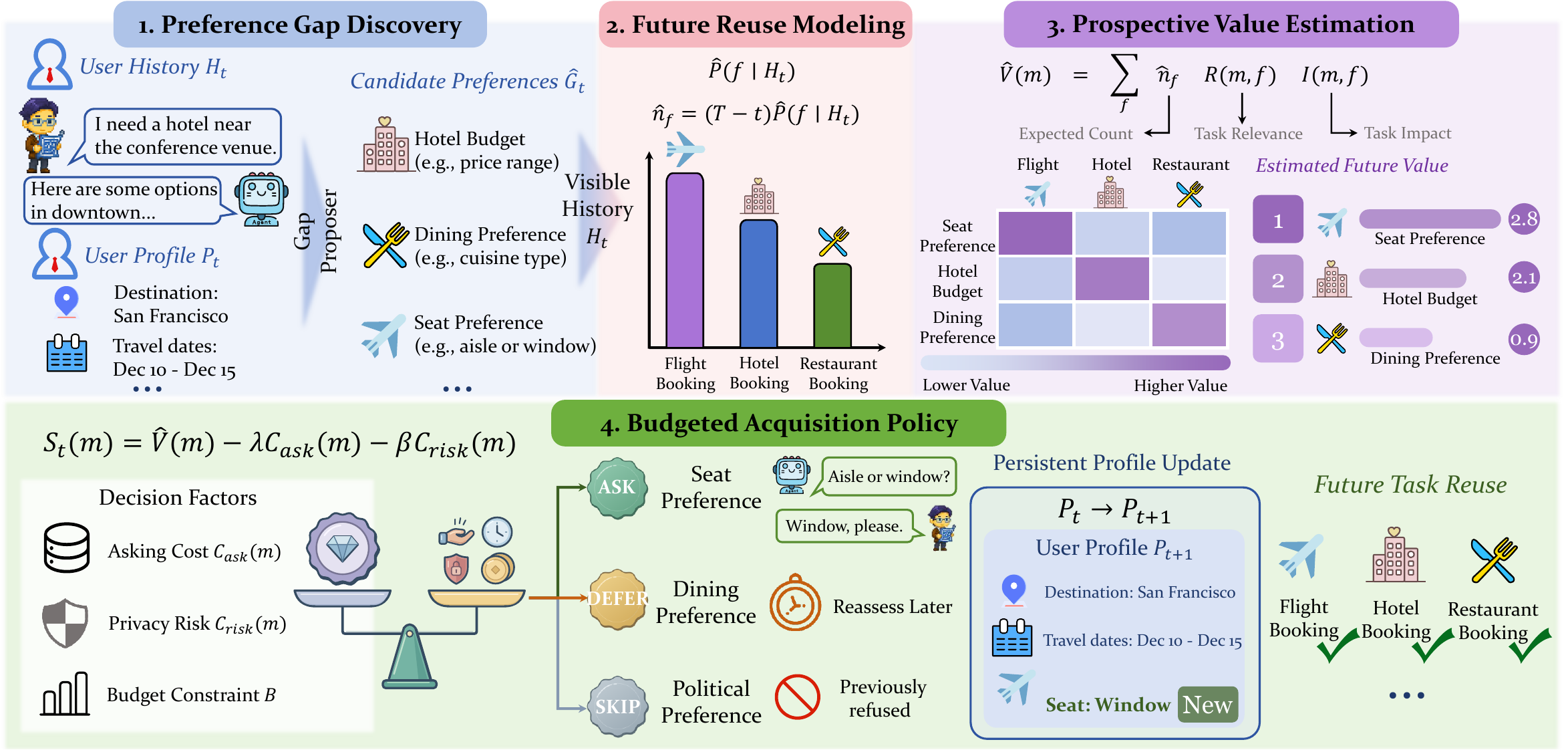}
\caption{Overview of \apa. \textcircled{1}~\emph{Gap discovery} proposes preferences $\mathcal{G}_t$ missing from the profile $P_t$. \textcircled{2}~\emph{Future reuse modeling} estimates how often each may be reused, from a coarse distribution over task families. \textcircled{3}~\emph{Prospective value estimation} scores each by relevance and impact. \textcircled{4}~\emph{Cost-aware acquisition} subtracts asking cost and disclosure risk and chooses \ask, \defer, or \skipa. Here, the seat preference is asked; the dining preference, rarely reused because restaurant tasks are rare, is deferred for later re-scoring, as is the hotel budget once the seat question has used this interaction's budget; and a previously refused political preference, which never enters the candidate set, is skipped. Answers update the profile for future tasks.}
\Description{Four-stage pipeline: gap discovery, future reuse modeling, a relevance-impact matrix for value estimation, and a cost-aware policy that asks about the seat preference, defers the rarely reused dining preference (and the hotel budget once the interaction budget is spent), and skips a previously refused political preference; answers update the persistent profile.}
\label{fig:pipeline}
\end{figure*}

\paragraph{Preference gaps.}
The \emph{gap} at step $t$ is the set of attributes the agent does not know:
\[
\mathcal{G}_t=\{\,m\in P^*_u\setminus P_t\,\}.
\]
We focus on attributes that are \emph{genuinely unobserved}: they cannot be extracted from $H_t$ or reliably inferred from it.
This excludes facts that were stated but not stored, which is the target of memory extraction~\cite{yang2026promem}, and facts that can be inferred from the dialogue, which is the target of profile inference~\cite{zhang2026personalagent}.

\paragraph{Utility gain and task relevance.}
The gain from knowing $m$ on task $\tau$ is
\[
\Delta U(\tau;m)=U(\tau\mid P_t\cup\{m\})-U(\tau\mid P_t).
\]
Let $\mathcal{R}(m)\subseteq\mathcal{F}$ be the task families in which $m$ can change the solver's action.
Then $\Delta U(\tau;m)=0$ whenever $f(\tau)\notin\mathcal{R}(m)$.
The agent can execute an available action without additional questions; clarification may still improve its quality; \apa separately values reuse after the current interaction.

\paragraph{Acquisition cost.}
Asking about $m$ incurs a fixed public per-attribute cost $C_{\text{ask}}(m)=c_m$, including on refusal.
Successfully disclosing $m$ also incurs disclosure risk $C_{\text{risk}}(m)$, determined by attribute sensitivity.
We write $C_\lambda^\beta(m)=\lambda C_{\text{ask}}(m)+\beta C_{\text{risk}}(m)$.
These are two separate terms, so no component of the cost is counted twice.

\paragraph{Prospective Personalization Value.}
Future tasks are drawn from a distribution conditioned on the history, $\tau_{t+1:T}\sim P(\cdot\mid H_t)$.
We model this distribution over task \emph{families}, not specific future events.
The value of acquiring $m$ at step $t$ is
\begin{equation}
\ppv_T(m)=\underbrace{\E_{\tau_{t+1:T}\sim P(\cdot\mid H_t)}\Big[\textstyle\sum_{k=t+1}^{T}\Delta U(\tau_k;m)\Big]}_{\text{expected personalization benefit}}-\;C_\lambda^\beta(m).
\label{eq:ppv}
\end{equation}
The cost is paid once, while the benefit accumulates over every future task in $\mathcal{R}(m)$.
\ppv thus evaluates whether an unknown preference is worth acquiring for reuse after the current interaction; improving the current task itself is left to clarification.

\paragraph{Decision.}
For each $m\in\mathcal{G}_t$ the agent chooses $a_m\in\{\ask,\defer,\skipa\}$.
\defer retains a recognized but unasked candidate for re-evaluation at a later acquisition opportunity; \skipa excludes a blocked candidate.
Questions asked in the same interaction share a budget $B$, so the agent selects
\begin{equation}
\begin{aligned}
S^\star=\arg\max_{S\subseteq\mathcal{G}_t}\;&\E\Big[\textstyle\sum_{k}\Delta U(\tau_k;S)\Big]-\textstyle\sum_{m\in S}C_\lambda^\beta(m)\\
\text{s.t.}\;&\textstyle\sum_{m\in S}C_{\text{ask}}(m)\le B.
\end{aligned}
\label{eq:budget}
\end{equation}
Here $\Delta U(\tau;S)$ is the joint gain of the set $S$.
It may be smaller than the sum of individual gains when attributes are correlated, for example a cabin class and a seat preference.
Equation~\eqref{eq:budget} defines a per-opportunity acquisition objective over the remaining tasks, rather than an optimal policy over future acquisition opportunities.

\paragraph{Current-task versus cross-task valuation.}
Current-task VoI~\cite{dong2026voi} scores $m$ at task $k$ by $g^{\mathrm{cur}}_k(m)-C_\lambda^\beta(m)$, where $g^{\mathrm{cur}}_k(m)$ is the expected gain of knowing $m$ on $\tau_k$ under the profile $P_k$ that the policy has reached, and it is applied anew at every task: the natural \emph{wait-until-needed} strategy. \ppv instead sums the \emph{fixed-profile} gains $g^{\mathrm{fix}}_{k\mid t}(m)=\E[\Delta U(\tau_k;m)]$ of Eq.~\eqref{eq:ppv}, all evaluated at the current profile $P_t$. The two coincide when other acquisitions leave the value of $m$ unchanged, but differ in general.
\begin{proposition}[What current-task valuation misses]
\label{prop:amortization}
\leavevmode\par
(i)~If $g^{\mathrm{cur}}_k(m)<C_\lambda^\beta(m)$ at every task $k$, in the decision states the policy actually reaches, then current-task VoI never acquires $m$, at any step and for any horizon.\par
(ii)~$\ppv_T(m)=\sum_{k=t+1}^{T}g^{\mathrm{fix}}_{k\mid t}(m)-C_\lambda^\beta(m)$, so $\ppv_T(m)>0$ if and only if $\sum_{k=t+1}^{T}g^{\mathrm{fix}}_{k\mid t}(m)>C_\lambda^\beta(m)$.\par
(iii)~If each relevant future task yields the same fixed-profile gain $g(m)>0$, then $\ppv_T(m)>0$ if and only if the expected number of relevant future tasks exceeds $C_\lambda^\beta(m)/g(m)$.
\end{proposition}
We call $m$ \emph{amortization-dependent} at step $t$ if $\max_k g^{\mathrm{fix}}_{k\mid t}(m)<C_\lambda^\beta(m)$ but $\sum_k g^{\mathrm{fix}}_{k\mid t}(m)>C_\lambda^\beta(m)$: at the current profile no single task justifies asking, yet reuse does.
By (ii), \ppv assigns such a preference positive net value, whether the agent then asks now or at the first relevant task; by (i), current-task VoI never acquires it unless other answers raise its current gain above the cost. The result is an elementary consequence of the definitions; its role is to define a diagnostic regime that we measure in \S\ref{sec:rq2}.
The proof is in Appendix~\ref{app:proof}.
\section{The \apa Framework}
\label{sec:method}

\apa estimates the terms of Eq.~\eqref{eq:ppv} with four modules (Figure~\ref{fig:pipeline}). Its core is counterfactual reuse valuation (\S\ref{sec:value}) combined with cost-sensitive budgeted selection (\S\ref{sec:policy}); gap discovery (\S\ref{sec:gap}) and future reuse modeling (\S\ref{sec:future}) are implementation components that supply candidates and reuse estimates.
To isolate the acquisition decision, all LLM roles share the same backbone, and all policies share the same deterministic task solver.
Prompts are given in Appendix~\ref{app:prompts}.

\subsection{Preference Gap Discovery}
\label{sec:gap}
The agent knows a complete public attribute catalogue but not the user\textquotesingle s missing values; it selects which unknown fields to evaluate for acquisition.
Given $(H_t,P_t)$ and the arrived task (null before the sequence), a proposer LLM selects $\min(K,|\mathcal A_t|)$ distinct IDs from the allowed unknown-field set $\mathcal A_t$, predicting stability and a natural-language question $q_i$. The catalogue supplies each description, so the resulting candidates retain the form $(m_i,\text{description}_i,\text{stability}_i,q_i)$.
The proposer must follow the following rules:
\begin{itemize}[leftmargin=*]
\item exclude attributes already in $P_t$ or stated in $H_t$, their publicly declared exact aliases, known refusals, and IDs already pending;
\item prefer reusable preferences, including constraints valid within the planning window, over transient details; infer stability from public descriptions, without gold labels;
\item never guess a value;

\end{itemize}
Candidate IDs are validated against the public catalogue; the evaluator computes gap recall against missing-field annotations (\S\ref{sec:analysis}).

\subsection{Future Reuse Modeling}
\label{sec:future}
\apa needs an estimate of how often each candidate may be reused. We instantiate this through a coarse distribution over future task families rather than predicting specific events, such as a particular conference trip.
It estimates a distribution over task families, $\hat P(f\mid H_t)$, and the expected number of occurrences of each family in the remaining horizon, $\hat n_f=(T-t)\,\hat P(f\mid H_t)$.
The LLM estimates nonnegative weights $w_f$ from visible history, profile and arrived task; code normalizes $\hat P(f)=w_f/\sum_g w_g$.
For example, a user who has booked two conference hotels is likely to book flights.

\subsection{Prospective Value Estimation}
\label{sec:value}
For each candidate $m$ and family $f$, code estimates relevance $R(m,f)\in[0,1]$ as the fraction of public prototypes whose active utility terms reference $m$, and impact $I(m,f)$ as their mean signed counterfactual utility increment.
$N_{\mathrm{proto}}=3$ prototypes per family (48 in total) are generated by the public task generator from the catalogue, independently of evaluated users and future sequences. Let $J(m,f)$ be the structurally relevant prototypes. Then $R=|J|/N_{\mathrm{proto}}$ and $I=|J|^{-1}\sum_{\tau\in J}G_q(m;\tau,P_t)$ (zero for empty $J$), where
\[
\begin{aligned}
G_q(m;\tau,P_t)&=\sum_z q_{P_t}(z)\,\Delta_\tau(m,z;P_t),\\
\Delta_\tau(m,z;P_t)&=U_\tau(D_\tau(P_t\cup\{m=z_m\}),z)\\
&\quad-U_\tau(D_\tau(P_t),z).
\end{aligned}
\]
Here $D_\tau$ is the shared solver and $q_{P_t}$ is a finite public belief over legal unknown-value states: categorical supports and representative numeric points, respecting coupled constraints (Appendix~\ref{app:impl}). Rebuilding this finite approximation after observing an answer can produce signed increments; structural relevance does not require a changed choice.
For option $o$, let $Q_\tau(o)\in[0,1]$ denote public option quality and $\phi_j(o,z)\in[0,1]$ its fit to soft preference $j$. The implemented task utility and shared decision rule are
\[
\begin{aligned}
s_\tau(o,z)&=w_qQ_\tau(o)+\sum_{j\in\mathcal S_\tau}w_j\phi_j(o,z),\\
U_\tau(o,z)&=\begin{cases}0,&\text{a hard constraint fails},\\0.2+0.8s_\tau(o,z),&\text{otherwise},\end{cases}\\
D_\tau(P)&=\arg\max_o\sum_zq_P(z)U_\tau(o,z).
\end{aligned}
\]
With soft terms, $w_q=0.3$ and the remaining $0.7$ is divided equally among those terms; otherwise $w_q=1$. A preference-inactive task has no preference constraints or fit terms. Ties are broken by option ID. Thus $G_q$ integrates the utility change of this fixed solver before and after an answer, rather than an LLM-assigned importance score.

The estimated benefit decomposes over task families:
\begin{equation}
\hat V(m)=\sum_{f\in\mathcal{F}}\hat n_f\,R(m,f)\,I(m,f).
\label{eq:vhat}
\end{equation}
This decomposition keeps how often a family occurs ($\hat n_f$) separate from how much $m$ matters within that family ($R\cdot I$).
The acquisition score subtracts the cost estimate:
$\mathrm{Score}(m)=\hat V(m)-\lambda\hat C_{\text{ask}}(m)-\beta\hat C_{\text{risk}}(m)$.
In the main experiments, costs come from the benchmark metadata.
The estimator approximates Eq.~\eqref{eq:ppv}, which counts only tasks $k>t$ that have not yet arrived. Within an opportunity, the family forecast is held fixed while answers trigger marginal-value recomputation. Stability is a proposal label, not a multiplier in $\hat V$.

\subsection{Budgeted Acquisition Policy}
\label{sec:policy}
We optimize Eq.~\eqref{eq:budget} with an adaptive greedy heuristic that selects by marginal score per unit cost (Algorithm~\ref{alg:greedy}); we do not claim an approximation guarantee, since monotonicity and submodularity of the objective are not established.
The value of a candidate is recomputed after each observed answer using the updated profile, which discounts redundant information.
Selection stops when no candidate has a positive marginal score or the budget is exhausted.
Recognized unasked attributes remain pending and are re-scored at later opportunities; known or refused fields are removed. Scores within floating-point tolerance of zero are treated as zero. Positive-value free questions have priority. Algorithm~\ref{alg:greedy} uses $\widehat{\mathcal G}_t\subseteq\mathcal G_t$, the recognized missing attributes that are not known to be unqueryable.

\begin{algorithm}[t]
\caption{Adaptive greedy acquisition heuristic at step $t$}
\label{alg:greedy}
\small
\begin{algorithmic}[1]
\State $S\gets\emptyset$, $b\gets B$
\While{an eligible affordable candidate remains}
  \For{$m\in\widehat{\mathcal G}_t\setminus S$}
    \State $\delta_m\gets{\hat V(m\mid P_t)}-\lambda\hat C_{\text{ask}}(m)-\beta\hat C_{\text{risk}}(m)$
  \EndFor
  \State $m^\star\gets\arg\max_{m:\,0<\hat C_{\text{ask}}(m)\le b}\;\delta_m/\hat C_{\text{ask}}(m)$
\State Prioritize positive-value free candidates
  \If{$m^\star$ is undefined \textbf{or} $\delta_{m^\star}\le0$} \textbf{break} \EndIf
  \State $S\gets S\cup\{m^\star\}$; $b\gets b-\hat C_{\text{ask}}(m^\star)$
\State $a_{m^\star}\gets\ask$; obtain $\mathrm{answer}(q_{m^\star})$
\State $P_t\gets P_t\cup\{(m^\star,\mathrm{answer}(q_{m^\star}))\}$ on a typed answer; otherwise record refusal permission
\EndWhile
\State Retain recognized unasked fields as \defer; omit known/refused fields
\State $P_{t+1}\gets P_t$
\end{algorithmic}
\end{algorithm}

\subsection{Profile Update and Reuse}
Answers are written as typed attribute--value pairs to a structured persistent profile; user-local permission is updated after success or refusal. Source, step and sensitivity are retained in the experiment logs.
Every later task in any domain can read them.
In the main comparison, all policies share the same simple profile store, so differences in results come from acquisition alone.
\section{\bench and Experimental Setup}
\label{sec:bench}

Existing long-term memory benchmarks either place the information needed for each question in the history or ask models to abstain when it is absent~\cite{wu2025longmemeval,maharana2024locomo}; none rewards acquiring missing information.
\apa instead requires preferences that are \emph{useful but initially absent from the history}.
It also requires task utilities that can be computed exactly, so that the benefit of a question can be measured without an LLM judge.
We therefore build \bench as a controlled longitudinal testbed.
The benchmarks discussed in \S\ref{sec:related} do not directly provide a protocol with hidden but queryable profiles and counterfactual utilities of unasked preferences. \bench is deliberately controlled so that the counterfactual utility of an unasked preference is known exactly, allowing us to measure whether each question was worth asking. This control comes at the cost of realism, which we discuss in \S\ref{sec:limitations}.

\subsection{Benchmark Construction}
\paragraph{Domains and task families.}
\bench covers five service domains: \emph{Travel} (flight booking, hotel selection, ground transport, conference trips), \emph{Dining} (restaurant choice, reservation, takeaway), \emph{Shopping} (product selection, gifts, repeat purchases), \emph{Scheduling} (meetings, recurring events, appointments), and \emph{Work/Conference} (conference, coordination, work events): 16 task families in total.
Task options are synthetic structured alternatives, rather than items taken from live Web catalogues.

\paragraph{Hidden user profiles.}
Each synthetic user has 27 attributes.
Each attribute has a ground-truth value and the following metadata:
\begin{itemize}[leftmargin=*]
\item \emph{stability}: stable, semi-stable, or ephemeral;
\item \emph{sensitivity}: low, medium, or high;
\item a normalized fixed asking cost $c_m\in[0,1]$;
\item the set of relevant task families $\mathcal{R}(m)$ and an impact level;
\item whether it is initially observed;
\item whether it is queryable.
\end{itemize}
Between 40\% and 60\% of useful attributes are hidden at $t=0$.
Profiles are stored separately from dialogues and are visible only to the user simulator and the evaluator.
The privileged oracle also accesses the hidden profile; practical policies receive the public catalogue and visible profile/history only.

\paragraph{Task sequences.}
For each user we save a separate initial context and 20 future tasks. Main CTU evaluates the first $T=10$ future tasks; horizon scaling uses nested prefixes $T\in\{1,3,5,10,20\}$.
The generator varies reuse frequency, cross-domain relevance, acquisition cost, sensitivity, and attribute dependence, to evaluate their effects on acquisition decisions. It includes five kinds of attributes:
\begin{itemize}[leftmargin=*]
\item attributes with heterogeneous reuse: some are used once, some repeatedly, some never;
\item cross-domain attributes (e.g., meeting mode) as well as domain-specific ones (e.g., seat preference);
\item \emph{distractors}: attributes that look personal but never affect utility;
\item attributes that are high-value but also high-cost or high-sensitivity;
\item correlated attributes, whose joint value may be less than the sum of their individual values.
\end{itemize}

\paragraph{Exact utilities and user simulator.}
As in evaluations that separate user-side private information from the agent~\cite{yao2025taubench,zhang2024agent4rec,balog2025usersim},
each task has a candidate set, for example flight options with seat, layover, and price attributes.
The solver picks the best candidate under the agent's current profile, and utility is the score of that pick under $P^*_u$.

The simulator answers only questions that map to a single queryable attribute and never reveals other attributes.
Requests for non-queryable or high-sensitivity fields are refused, and the refused question still incurs its asking cost. Risk is charged only on successful disclosure.
The main experiments use a deterministic structured answer mode.

\paragraph{Splits.}
The two disjoint test splits contain IID users and users with unseen preference combinations. A separate amortization-enriched test split of 60 users (\S\ref{sec:rq2}) was added afterwards without changing these splits.
All splits were frozen before any test-set evaluation. The train split is not used by any policy, for parameter selection, or for prototype sampling; it is retained as a benchmark split and used to verify that the unseen combinations occur in neither train nor dev.
Table~\ref{tab:stats} gives statistics.

\begin{table*}[tp]
\centering
\caption{\bench statistics by primary-user domain. Test combines 90 IID + 90 unseen users; each cell aggregates its 36 test users unless train/dev counts are shown. Attribute inventories are shared across domains; hidden fractions use all 27 fields. Structural use is counted over the first $T=10$ future tasks.}
\label{tab:stats}
\fontsize{7.8}{9.2}\selectfont\setlength{\tabcolsep}{3pt}
\begin{tabular*}{\textwidth}{@{\extracolsep{\fill}}lrrrrr}
\toprule
Statistic & Travel & Dining & Shopping & Scheduling & Work \\
\midrule
\# users (train/dev/test) & 18/18/36 & 18/18/36 & 18/18/36 & 18/18/36 & 18/18/36 \\
\# task families & 4 & 3 & 3 & 3 & 3 \\
\# attributes / user & 27 & 27 & 27 & 27 & 27 \\
\% hidden at $t=0$ & 53.1 & 53.1 & 53.1 & 53.1 & 53.1 \\
\% high-sensitivity & 11.1 & 11.1 & 11.1 & 11.1 & 11.1 \\
mean uses / used field & 1.687 & 2.000 & 1.862 & 1.780 & 2.178 \\
\bottomrule\end{tabular*}
\end{table*}

\subsection{Baselines}
\label{sec:setup}
All policies share the task solver, profile store and simulator; every LLM role uses the same backbone. The main table compares the core measured policies; the implementation also includes the simple policies listed below.
\begin{itemize}[leftmargin=*]
\item \textbf{Controls.} \emph{Ask-Nothing} uses only the initially observed profile. \emph{Ask-Everything} asks all unknown publicly eligible fields without a binding budget and pays for every attempted question, including refusals.
\item \textbf{Simple policies.} Selection obeys the acquisition-cost cap.
\begin{itemize}
\item \emph{Random-$k$}: random candidates, averaged over 5 seeds.
\item \emph{Frequency}: estimated reuse only.
\item \emph{Impact-only}: ranks candidates by $\max_{f\in\mathcal F} I(m,f)$, the largest family-specific counterfactual impact, without reuse weighting or cost/risk penalties in the ranking.
\item \emph{LLM-Importance}: the same backbone ranks ``which unknown facts are most worth knowing'', without task decomposition.
\end{itemize}
\item \textbf{Current-task VoI}~\cite{dong2026voi}: an adaptation closer to the LLM-based current-task VoI formulation of Dong et al. (ACL 2026), in which Qwen predicts relative probabilities of unknown preference states for the current task. The shared solver computes the resulting current-task information gain; no future task is used. At the default setting it reaches NU 6.480 (CTU 6.640, AC 0.318, 1.82 questions), below the public-belief version, which we therefore use as the stronger current-task baseline.
\item \textbf{Current-task VoI (public)}: a current-task rule using the same public finite belief over unknown preference states as \apa, with the same solver, cost/risk coefficients and positive-score selection. It has no access to future tasks.
\item \textbf{Next-Task Only}: public-belief value for the actual next task instance, without hidden answers; this is privileged lookahead.
\item \textbf{Oracle Future-VoI}: uses exact chronological subset dynamic programming to jointly optimize acquisitions and their timing, maximizing NU at the evaluation prices under the budget, with access to hidden answers and the actual future task sequence. It is a privileged reference, not a practical competing method; oracle regret is measured relative to it. Its dynamic program and tie-breaking are specified in Appendix~\ref{app:impl}.
\item \textbf{Oracle Future-VoI (unlimited)}: Oracle Future-VoI without a budget limit.
\item \textbf{Cost-matched heuristics.} Random-$k$, Frequency, Impact-only and LLM-Importance are all rerun under each user's realized \apa acquisition budget, $C^{\text{total}}_{\text{baseline},u}\le C^{\text{total}}_{\text{\apa},u}$, in addition to the per-interaction cap $B$. This separates acquisition quality from resource use.
\end{itemize}
We compare \apa with current-task VoI not because the two solve identical decisions, but because current-task VoI represents the natural wait-until-needed strategy. The comparison tests whether prospective acquisition provides value beyond deferring every question until it affects the current task. This comparison evaluates cross-task against current-task valuation under their respective acquisition protocols, including \apa's additional pre-sequence opportunity; it does not separately identify the effect of acquisition timing.
Every policy is run with the same per-interaction budget $B$, except Ask-Everything and Oracle Future-VoI (unlimited).
The cap resets at each arrival. Prospective policies also receive one pre-sequence opportunity with no current task; current-task policies do not.
Pareto sweeps vary $B$ and $\lambda$ for every policy.

\subsection{Metrics and Implementation}
\paragraph{Metrics.}
Benefit is measured as \emph{cumulative task utility}, $\text{CTU}=\sum_t U(\tau_t)$.
Cost is measured as \emph{acquisition cost} AC $=\sum C_{\text{ask}}$ and \emph{risk cost} RC $=\sum C_{\text{risk}}$; we also report the raw number of questions.
Our primary metric is \emph{net utility}, NU $=\text{CTU}-\lambda\,\text{AC}-\beta\,\text{RC}$.
We further report: utility per question relative to Ask-Nothing; oracle regret ($\text{NU}_{\text{oracle}}-\text{NU}$); \emph{unnecessary query rate}, the fraction of asked attributes with zero realized utility within the horizon; \emph{missed valuable rate}; gap precision and recall; and the rank agreement between predicted and oracle value (NDCG~\cite{jarvelin2002cumulated} and Spearman correlation).

\paragraph{Statistics.}
The resampling unit is a user's task sequence.
We report means with 95\% pointwise percentile bootstrap confidence intervals~\cite{efron1994introduction} and use two-sided centered paired bootstrap tests (2,000 resamples, plus-one correction) with Holm correction within separate families: the 15 main-comparison contrasts (five comparators on IID, unseen and IID + unseen), the four cost-matched contrasts, the two amortization-cohort contrasts, the 15 ablation contrasts, the 15 horizon contrasts, the three endpoint-growth tests, and, on the amortization-enriched set, its eight non-oracle contrasts and five horizon contrasts (Table~\ref{tab:tests}).

\paragraph{Implementation.}
All LLM roles use Qwen3-8B~\cite{yang2025qwen3}, temperature 0 and thinking disabled, for candidate recognition, forecasting and LLM baselines. Value integration and task solving are deterministic. The exact prompts are listed in Appendix~\ref{app:prompts}.
Up to $K=10$ fresh candidates are proposed per opportunity; recognized pending candidates are retained separately.
All model outputs are cached, and every table and figure is generated from the logged results.
Defaults are $B=0.5$, $\lambda=0.5$, $\beta=0.25$, $T=10$, $K=10$. All reported Overall values pool IID and unseen users. A development-set comparison supports $B=0.5$: lowering it from 1 to 0.5 changes CTU by $-0.007$, increases NU by 0.034 and reduces oracle regret ($B=1$: CTU 7.074, NU 6.681, regret 0.336; $B=0.5$: CTU 7.067, NU 6.715, regret 0.298, over 90 development users); test results at $B=1$ are reported in Appendix~\ref{app:allpolicies}.
\section{Results}
\label{sec:results}

We organize the results around four research questions.

\subsection{RQ1: Does active acquisition improve long-term personalization?}
Current-task valuation can acquire many useful preferences once a relevant task arrives, so we expect \apa's overall advantage over it to be modest and to concentrate on preferences that only cumulative reuse justifies, which RQ2 examines.

\begin{table*}[tp]
\centering
\caption{Main results on \bench: Overall = 90 IID + 90 unseen users, $T=10$, $B=0.5$, $\lambda=0.5$, $\beta=0.25$, $K=10$. CTU: cumulative task utility; AC/RC: asking/disclosure risk cost; NU: net utility; \#Q: questions; Unn.: unnecessary query rate. Cost-matched rows obey both the per-interaction budget and a per-user cap on total asking cost (Appendix~\ref{app:impl}); Random-$k$ averages five seeds. Pointwise 95\% CIs appear in Table~\ref{tab:overallci}; paired tests are in Table~\ref{tab:tests}.}
\label{tab:main}
\fontsize{8}{9.4}\selectfont\setlength{\tabcolsep}{2.5pt}
\begin{tabular*}{\textwidth}{@{\extracolsep{\fill}}lrrrrrrr}
\toprule
Policy & CTU$\uparrow$ & AC$\downarrow$ & RC$\downarrow$ & NU$\uparrow$ & \#Q$\downarrow$ & Unn.$\downarrow$ & Regret$\downarrow$ \\
\midrule
Ask-Nothing & 6.289 & 0.000 & 0.000 & 6.289 & 0.000 & -- & 0.769 \\
Ask-Everything & 7.263 & 3.977 & 1.428 & 4.918 & 13.472 & 0.728 & 2.147 \\
LLM-Importance & 7.080 & 2.455 & 0.883 & 5.631 & 11.294 & 0.741 & 1.427 \\
\midrule
\multicolumn{8}{l}{\emph{Heuristics with per-user asking cost capped at \apa's realized expenditure}}\\
Random-$k$ (cost-matched) & 6.585 & 0.654 & 0.096 & 6.234 & 3.696 & 0.698 & 0.824 \\
Frequency (cost-matched) & 6.748 & 0.659 & 0.072 & 6.400 & 3.783 & 0.592 & 0.658 \\
Impact-only (cost-matched) & 7.037 & 0.666 & 0.083 & 6.683 & 3.744 & 0.462 & 0.376 \\
LLM-Importance (cost-matched) & 6.603 & 0.660 & 0.339 & 6.188 & 3.011 & 0.704 & 0.870 \\
\midrule
\multicolumn{8}{l}{\emph{Value-based policies, per-interaction budget only}}\\
Current-task VoI (public) & 6.853 & 0.379 & 0.000 & 6.664 & 2.044 & 0.314 & 0.395 \\
\apa (ours) & 7.109 & 0.680 & 0.058 & 6.755 & 3.717 & 0.387 & 0.304 \\
Oracle Future-VoI & 7.265 & 0.400 & 0.028 & 7.058 & 2.122 & 0.029 & 0.000 \\
\bottomrule\end{tabular*}
\par\smallskip\begin{minipage}{\textwidth}\fontsize{7.4}{8.7}\selectfont Regret is relative to Oracle Future-VoI, except for Ask-Everything, whose regret is relative to Oracle Future-VoI (unlimited). The full-policy comparison at $B=1$ is in Appendix~\ref{app:allpolicies}, Table~\ref{tab:allpolicies}.\end{minipage}
\end{table*}

Table~\ref{tab:main} reports the main comparison.
\apa achieves the highest mean net utility among the measured practical core policies, at 6.755, compared with 6.664 for the strongest baseline, Current-task VoI (public) (paired difference 0.091; 95\% CI [0.019, 0.174]; Holm $p=0.0615$).
Ask-Everything attains the highest raw CTU among these practical policies, but asks 3.62 times as many questions.
After costs, its net utility falls below that of \apa.
The comparison with exhaustive asking supports selective acquisition: \apa retains most of the utility gain while asking fewer questions. Results for all twelve policies at $B=1$ are in Appendix~\ref{app:allpolicies} (Table~\ref{tab:allpolicies}).

Cost-matched controls (Table~\ref{tab:main}; Appendix Table~\ref{tab:costmatched}) rerun the heuristics with each user's asking cost capped at \apa's realized expenditure at $B=0.5$ (mean AC 0.654--0.666 versus 0.680). Under matched cost caps, \apa's NU (6.755) exceeds that of Random-$k$ (6.234), Frequency (6.400), and LLM-Importance (6.188) (Random-$k$: $\Delta$NU $=0.521$, 95\% CI [0.425, 0.635], Holm $p=0.0020$; Frequency: $\Delta$NU $=0.355$, 95\% CI [0.280, 0.435], Holm $p=0.0020$; LLM-Importance: $\Delta$NU $=0.567$, 95\% CI [0.481, 0.655], Holm $p=0.0020$), so \apa achieves higher NU at comparable asking expenditure. Impact-only comes close (6.683; $\Delta$NU $=0.072$, 95\% CI [0.031, 0.114], Holm $p=0.0025$): the gap remains significant under the same per-user spending cap. This shows that \apa outperforms the Impact-only heuristic implemented here, which also differs from \apa in family aggregation, cost terms and selection rule; it is not a pure ablation of reuse weighting. The tests pair 180 users, use 2,000 bootstrap replicates with global resampling of the five Random-$k$ seeds, and apply Holm correction jointly to these four NU contrasts.

\subsection{RQ2: What does current-task valuation miss?}
\label{sec:rq2}

Proposition~\ref{prop:amortization} predicts a regime in which current-task valuation cannot acquire valuable preferences. We first ask whether this regime occurs at meaningful scale. We classify every (user, initially unknown, queryable, non-high-sensitivity attribute) pair in the test set by its per-task gains and weighted asking cost. Here $g^{\mathrm{fix}}_{k\mid 0}(m)=G_q(m;\tau_k,P_0)$ is the fixed-profile gain of Proposition~\ref{prop:amortization} on saved future task $\tau_k$, under the same public belief that Current-task VoI (public) uses and the initial visible profile $P_0$; we write $g_k$ for it below, and $C=\lambda c_m+\beta C_{\text{risk}}(m)$ with the evaluation prices $\lambda=0.5$ and $\beta=0.25$. We call this the \emph{initial-profile amortization diagnostic}: because it uses fixed-profile gains at $P_0$ rather than current gains along the trajectory, Proposition~\ref{prop:amortization} does not guarantee that Current-task VoI (public) acquires none of these pairs, and we report its acquisition rate as an observation. Appendix~\ref{app:proof} explains the observed acquisitions of diagnostic amortization-dependent attributes by Current-task VoI (public). The classes are \emph{single-use justified} ($\max_k g_k\ge C$), \emph{amortization-dependent} ($\max_k g_k<C$ but $\sum_k g_k>C$), and \emph{not worth acquiring} ($\sum_k g_k\le C$). We call a pair a \emph{budget-feasible valuable missing preference} if $\sum_k g_k>C$ and its asking cost fits the per-interaction budget, $c_m\le B$; pairs with $c_m>B$ cannot be asked by a budgeted policy and are excluded. Six attributes have $c_m>B$ and are excluded; two of them would otherwise be amortization-dependent in 26 pairs at $T=20$.
At $T=20$ and $B=0.5$, 198 pairs across 120 users of the original test set (63 IID, 57 unseen) are amortization-dependent: 198/576, or $34.4\%$, of budget-feasible valuable missing preferences. At $T=10$, the corresponding figures are 91 pairs across 71 users (36 IID, 35 unseen), 91/430 or $21.2\%$; 2 additional amortization-dependent pairs are excluded for exceeding $B$.
Current-task VoI (public) acquires 10/198 ($5.1\%$) of these pairs at $T=20$ and 3/91 ($3.3\%$) at $T=10$.
\apa acquires $70.3\%$ (64/91) at $T=10$ and $80.3\%$ (159/198) at $T=20$; Oracle Future-VoI acquires $42.9\%$ (39/91) and $44.9\%$ (89/198), respectively (Table~\ref{tab:amortization}). The oracle optimizes joint net utility, rather than acquisition rate.
Oracle Future-VoI acquires none of these pairs before their first use (Table~\ref{tab:amortization}). Because its dynamic program breaks ties in favor of deferral (Appendix~\ref{app:impl}), this shows that an NU-optimal schedule within the oracle's candidate pool never needs to acquire them earlier, not that waiting is strictly better; what separates it from Current-task VoI (public) is that it prices cross-task reuse, not that it asks earlier.
\apa acquires more of these pairs than the oracle (80.3\% versus 44.9\% at $T=20$) but realizes less attributed gain (0.307 versus 0.369), consistent with less precise selection or over-acquisition within this class, and with the modest rank agreement of its value estimates (Spearman 0.296 on this $T=20$ cohort; Appendix~\ref{app:calib}). A higher acquisition rate is therefore not itself the goal: the NU-maximizing oracle acquires fewer than half of these preferences, so the task is to select those whose reuse repays their cost, and \apa's lower attributed gain leaves room for more precise selection. \apa both estimates cross-task value and decides what to acquire under cost and budget; its gap to the oracle shows that, even with cross-task benefit counted, valuation and selection remain imprecise. Because the oracle also knows the hidden answers and the actual future, our experiments cannot attribute this gap to forecasting, value estimation, greedy selection or timing. Attributed gain follows the actual order of questions: for each designated acquisition, it is the signed change in realized CTU on the arrived and remaining tasks when that answer is added to the answers obtained before it. It is gross of asking cost, summed over designated acquisitions, and averaged over all cohort users, with users who acquire none counted as zero; because it conditions on earlier answers, it differs from removing one attribute from the final profile.
Among users with at least one designated pair, \apa's improvement in whole-sequence NU over Current-task VoI (public) (paired mean differences 0.148 at $T=10$ and 0.263 at $T=20$) is significant at $T=20$ (raw $p=0.010$, Holm $p=0.021$ over the two horizons) but not at $T=10$ ($p=0.074$) (Tables~\ref{tab:amortizationresults} and~\ref{tab:tests}). The tests pair the same 120 users at $T=20$ and 71 at $T=10$, each with at least one feasible amortization-dependent pair; they use 2,000 two-sided centered user-bootstrap resamples with a plus-one correction. Holm correction covers these two NU contrasts, not attribute pairs or the attributed gains in Table~\ref{tab:amortization}. These results support an advantage for these users, but do not attribute the whole gain to the designated attributes.
\begin{table}[!htb]
\centering
\caption{Amortization-enriched test set (60 users, $T=10$, $B=0.5$): NU, paired difference \apa minus each policy with Holm $p$ over the eight non-oracle baselines, and acquisition of its 202 amortization-dependent pairs.}
\label{tab:amorttest}
\fontsize{8}{9.4}\selectfont\setlength{\tabcolsep}{2.5pt}
\begin{tabular*}{\columnwidth}{@{\extracolsep{\fill}}lrrrr}
\toprule
Policy & NU$\uparrow$ & $\Delta$NU & $p_{\rm Holm}$ & Acq. \\
\midrule
Ask-Nothing & 5.706 & +0.679 & 0.004 & 0\% \\
Ask-Everything & 4.534 & +1.852 & 0.004 & -- \\
LLM-Importance & 5.347 & +1.039 & 0.004 & -- \\
Random-$k$ (cost-matched) & 5.815 & +0.570 & 0.004 & -- \\
Frequency (cost-matched) & 6.020 & +0.365 & 0.004 & -- \\
Impact-only (cost-matched) & 6.271 & +0.114 & 0.109 & -- \\
LLM-Importance (cost-matched) & 5.643 & +0.743 & 0.004 & -- \\
Current-task VoI (public) & 6.032 & +0.354 & 0.004 & 0.0\% \\
\apa (ours) & 6.385 & -- & -- & 73.8\% \\
Oracle Future-VoI & 6.902 & $-0.517$ & -- & 41.1\% \\
\bottomrule\end{tabular*}
\end{table}

\paragraph{An amortization-enriched test set.}
To test the mechanism where it matters most, we generated 16,560 users with the unchanged generator and schemas and kept, for the IID and the unseen-combination schema separately, the 30 users with the most amortization-dependent pairs under the initial-profile diagnostic, without using any policy outcome. Generation stopped once the 60 users held at least 200 such pairs at $T=10$; they hold 202 (3, 4 and 5 pairs for 40, 18 and 2 users). It contains 210 such pairs at $T=20$. It is a mechanism diagnostic, not a natural distribution: its users' primary domains are skewed toward dining (40 of 60). Appendix~\ref{app:amorttest} gives the horizon contrasts and the APA-v2 variants.
On this set (Table~\ref{tab:amorttest}), Current-task VoI (public) acquires none of the 202 pairs, whereas \apa acquires 73.8\% and the oracle 41.1\%. \apa's NU exceeds that of Current-task VoI (public) by 0.354 (95\% CI [0.194, 0.533]; Holm $p=0.004$) and that of every other non-oracle baseline except cost-matched Impact-only ($+0.114$, [$-0.011$, 0.270], Holm $p=0.109$).
Conversely, \apa does not neglect preferences that a single task justifies: on the cohorts above (at $T=20$) it acquires 241 of 242 such pairs at $T=20$, against 234 for Current-task VoI (public). Consistent with this, a variant that adds a current-task value term to \apa's score (APA-v2; formula and per-horizon results in Appendix~\ref{app:amorttest}) shows no consistent NU gain across horizons on either test set.
Under the initial profile, preferences that need cumulative reuse to cover their asking cost thus make up a substantial share of valuable missing preferences; Current-task VoI (public) rarely acquires them, and the resulting advantage of cross-task valuation appears at the longer horizon.

\begin{table*}[tp]
\centering
\caption{Initial-profile amortization diagnostic: feasible pairs with $c_m\le B=0.5$, $\lambda=0.5$, $\beta=0.25$, $K=10$. Overall cohorts contain at least one such pair. Rates divide by all designated pairs, including unacquired pairs; counts appear in parentheses.}
\label{tab:amortization}
\fontsize{8}{9.5}\selectfont\setlength{\tabcolsep}{3pt}
\begin{tabular*}{\textwidth}{@{\extracolsep{\fill}}lrrrrrr}
\toprule
& \multicolumn{3}{c}{$T=10$: 91 pairs, 71 users (36 IID + 35 unseen)} & \multicolumn{3}{c}{$T=20$: 198 pairs, 120 users (63 IID + 57 unseen)} \\
\cmidrule(lr){2-4}\cmidrule(lr){5-7}
Policy & Acquired & Before use & Attr. gain & Acquired & Before use & Attr. gain \\
\midrule
Ask-Nothing & 0.0\% (0) & 0.0\% (0) & 0.000 & 0.0\% (0) & 0.0\% (0) & 0.000 \\
Ask-Everything & 100.0\% (91) & 100.0\% (91) & 0.254 & 100.0\% (198) & 100.0\% (198) & 0.576 \\
LLM-Importance & 100.0\% (91) & 35.2\% (32) & 0.204 & 100.0\% (198) & 49.0\% (97) & 0.431 \\
Current-task VoI (public) & 3.3\% (3) & 0.0\% (0) & -0.001 & 5.1\% (10) & 0.0\% (0) & 0.017 \\
\apa & 70.3\% (64) & 33.0\% (30) & 0.227 & 80.3\% (159) & 46.5\% (92) & 0.307 \\
Oracle Future-VoI & 42.9\% (39) & 0.0\% (0) & 0.279 & 44.9\% (89) & 0.0\% (0) & 0.369 \\
\bottomrule\end{tabular*}
\par\smallskip\begin{minipage}{\textwidth}\fontsize{7.4}{8.7}\selectfont
Before use means acquired strictly before the arrival of the first future task whose active utility terms reference that field; acquisition upon that arrival is not counted. Attributed gain is the signed realized CTU change over the arrived and remaining tasks when a designated answer is added, conditioning on all earlier answers, summed over designated acquisitions and averaged over all cohort users (zero if none). This chronological attribution is gross utility, not NU or a separate significance test. Oracle Future-VoI breaks ties in favor of deferral (Appendix~\ref{app:impl}); Ask-Everything has no budget cap.
\end{minipage}
\end{table*}

\begin{table*}[tp]
\centering
\caption{Whole-sequence outcomes on the budget-feasible initial-profile amortization cohorts: each user has at least one designated pair. $B=0.5$, $\lambda=0.5$, $\beta=0.25$, $K=10$. MVMR is the missed valuable rate (Appendix~\ref{app:impl}); $\Delta$CTU is the paired-cohort mean gain over Ask-Nothing.}
\label{tab:amortizationresults}
\fontsize{8}{9.5}\selectfont\setlength{\tabcolsep}{3pt}
\begin{tabular*}{\textwidth}{@{\extracolsep{\fill}}lrrrrrr}
\toprule
& \multicolumn{3}{c}{$T=10$: 71 users (36 IID + 35 unseen)} & \multicolumn{3}{c}{$T=20$: 120 users (63 IID + 57 unseen)} \\
\cmidrule(lr){2-4}\cmidrule(lr){5-7}
Policy & MVMR$\downarrow$ & $\Delta$CTU$\uparrow$ & NU$\uparrow$ & MVMR$\downarrow$ & $\Delta$CTU$\uparrow$ & NU$\uparrow$ \\
\midrule
Ask-Nothing & 1.000 & 0.000 & 6.291 & 1.000 & 0.000 & 12.437 \\
Ask-Everything & 0.000 & 1.044 & 4.988 & 0.000 & 2.098 & 12.153 \\
LLM-Importance & 0.027 & 0.903 & 5.726 & 0.041 & 1.864 & 12.817 \\
Current-task VoI (public) & 0.459 & 0.516 & 6.639 & 0.463 & 1.231 & 13.462 \\
\apa & 0.188 & 0.868 & 6.787 & 0.150 & 1.862 & 13.725 \\
Oracle Future-VoI & 0.109 & 1.068 & 7.138 & 0.167 & 2.109 & 14.267 \\
\bottomrule\end{tabular*}
\par\smallskip\begin{minipage}{\textwidth}\fontsize{7.4}{8.7}\selectfont
Outcomes cover all tasks in each cohort user's sequence, not only designated attributes. Undefined MVMR denominators are omitted: 49 users contribute at $T=10$ and 102 at $T=20$ for every policy. Ask-Everything has no budget cap.
\end{minipage}
\end{table*}

\subsection{RQ3: Is selective acquisition more efficient than exhaustive asking?}

\begin{figure*}[tp]
\centering
\includegraphics[width=\textwidth]{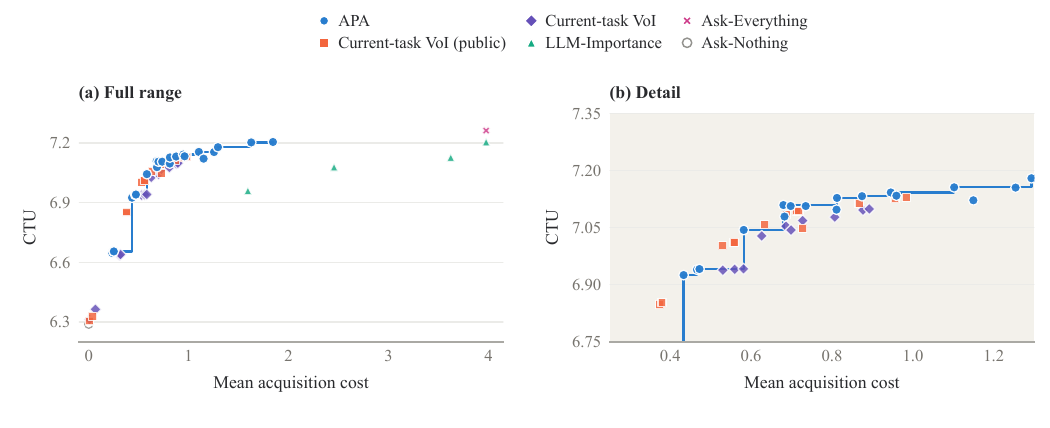}
\caption{Acquisition cost versus CTU: full range (a) and detail (b). Overall = 90 IID + 90 unseen users, $T=10$, $K=10$, $\beta=0.25$; all $B\in\{0.25,0.5,0.75,1\}$ and selection $\lambda\in\{0,0.1,0.25,0.5,1,2\}$ settings. Points retain the saved mean coordinates; the APA step line is its observed configuration envelope.}
\Description{Two side-by-side panels show acquisition cost versus CTU, with the full range at left and a magnified central region at right.}
\label{fig:pareto}
\label{fig:paretoctuzoom}
\end{figure*}

\begin{figure*}[tp]
\centering
\includegraphics[width=\textwidth]{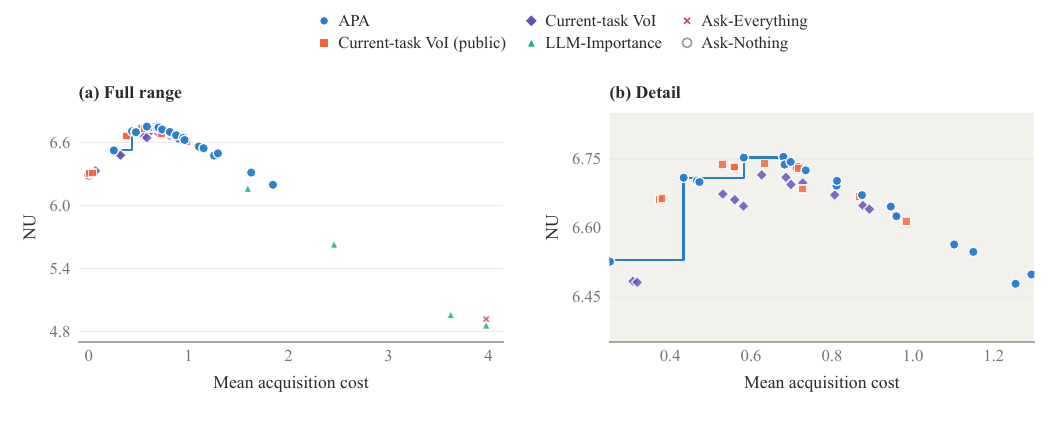}
\caption{Acquisition cost versus NU: full range (a) and detail (b). Overall = 90 IID + 90 unseen users, $T=10$, $K=10$, $\beta=0.25$; all $B\in\{0.25,0.5,0.75,1\}$ and selection $\lambda\in\{0,0.1,0.25,0.5,1,2\}$ settings. Points retain the saved mean coordinates; the APA step line is its observed configuration envelope. NU uses common evaluation prices $\lambda=0.5$, $\beta=0.25$.}
\Description{Two side-by-side panels show acquisition cost versus NU, with the full range at left and a magnified central region at right.}
\label{fig:paretonu}
\label{fig:paretonuzoom}
\end{figure*}

Figures~\ref{fig:pareto} and~\ref{fig:paretonu} sweep the acquisition budget and selection penalty for CTU and NU, respectively; each shows the full range and a detailed view.
\apa reaches 84.2\% of Ask-Everything\textquotesingle s utility gain over Ask-Nothing using 27.6\% of its questions at the default setting.
Across the saved sweep, \apa\textquotesingle s observed CTU envelope exceeds the best feasible practical-baseline configurations over representative mean-cost ceilings $[0.680,1.844)$; the corresponding question-count range is $[3.717,7.550)$. For NU the respective ranges are $[0.582,1.844)$ and $[3.284,7.550)$. The comparison includes Current-task VoI and Current-task VoI (public), Ask-Nothing, Ask-Everything and LLM-Importance.
These finite-grid mean envelopes describe the observed trade-off; they do not establish simultaneous statistical dominance. The detailed views make closely spaced configurations visible without changing their coordinates.

\FloatBarrier

\subsection{RQ4: Does the benefit grow with the horizon?}

\begin{figure}[tp]
\centering\includegraphics[width=\columnwidth]{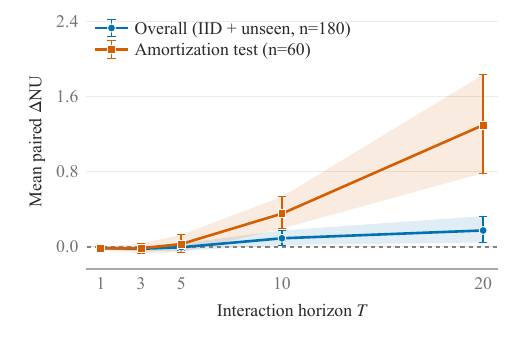}
\caption{Horizon scaling: paired NU(APA) minus NU(Current-task VoI (public)) on the IID + unseen test set (180 users) and the amortization-enriched test set (60 users); $B=0.5$, $\lambda=0.5$, $\beta=0.25$, $K=10$, nested saved prefixes. Bands and error bars are pointwise 95\% paired-user bootstrap CIs.}
\Description{Paired net-utility differences over five horizons.}
\label{fig:horizon}
\end{figure}

Figure~\ref{fig:horizon} varies $T$.
At $T=1$ their mean NU difference is $-0.011$ rather than zero. The implemented \apa values only tasks that have not yet arrived (\S\ref{sec:value}): with a single task, it can act only before the task arrives, whereas Current-task VoI (public) can still ask about that task once it is known.
For \apa, each $T$ uses a matching valuation horizon, decreasing with the tasks still to arrive, so longer horizons leave more reuse to amortize the one-time asking cost.
On the original IID + unseen test set, over longer interaction prefixes from $T=1$ to $T=20$, the mean advantage increases from $-0.011$ to 0.175, consistent with the amortization hypothesis; the increase is not monotone at short horizons (Table~\ref{tab:horizon}).
Individual horizon contrasts are Holm-corrected across all 15 horizon tests, so the $T=10$ contrast has a larger adjusted $p$ here than in the main-comparison family of RQ1. Paired endpoint growth is 0.186 (95\% CI [0.057, 0.339]; Holm $p=0.0330$ in the three endpoint-growth tests).
On the amortization-enriched set the advantage is $-0.017$ at $T=1$, 0.354 at $T=10$ and 1.296 at $T=20$ (95\% CI [0.786, 1.834]), significant at $T=10$ and $T=20$ (Holm $p=0.0025$ over five horizons) and not at $T\le 5$ (Table~\ref{tab:amorthorizon}); we did not test its growth across horizons directly on this set.
\section{Analysis}
\label{sec:analysis}

\begin{table}[tp]
\centering
\caption{Ablations: Overall 180 users (IID + unseen), $T=10$, $B=0.5$, $K=10$. NU always uses evaluation $\lambda=0.5$, $\beta=0.25$; No Cost/No Risk change only the selection coefficient. No FutureDist uses uniform task-family weights. Next-Task Only values the actual next task instance, a privileged lookahead without hidden answers. Gold Candidate Schema bypasses $K$ with the complete publicly eligible missing-field pool.}
\label{tab:ablation}
\fontsize{8.2}{9.6}\selectfont\setlength{\tabcolsep}{3pt}
\begin{tabular*}{\columnwidth}{@{\extracolsep{\fill}}lrr}
\toprule
Variant & NU$\uparrow$ & Unn.$\downarrow$ \\
\midrule
\apa (full) & 6.755 & 0.387 \\
No FutureDist & 6.760 & 0.342 \\
Next-Task Only & 6.683 & 0.315 \\
No Cost ($\lambda=0$) & 6.498 & 0.557 \\
No Risk ($\beta=0$) & 6.727 & 0.397 \\
Gold Candidate Schema & 6.753 & 0.378 \\
\bottomrule\end{tabular*}
\end{table}

\paragraph{Observation 1: Modeling cost prevents over-asking; modeling risk changes what is asked.}
Without the cost term, CTU rises by 0.070, but the number of questions grows to 1.79 times the original count and common-price NU drops by 0.257 (Holm $p=0.0075$).
Without the risk term, successful disclosure risk rises from 0.058 to 0.092, while CTU changes by 0.007. The NU contrast has Holm $p=0.0660$; high-sensitivity requests are refused in every variant.

\paragraph{Observation 2: Supplying the complete candidate catalogue has little effect on mean NU.}
The proposer reaches gap recall of 0.500 and precision of 1.000 against the evaluator\textquotesingle s missing-field set. Precision is constrained by catalogue validation, rather than human semantic assessment.
Supplying every publicly eligible missing-field ID changes NU from 6.755 to 6.753, without reducing oracle regret. It supplies no hidden answers or gold stability labels; this result does not isolate all remaining error sources.

\paragraph{Observation 3: Acquisition rates differ across attribute classes.}
\begin{figure}[tp]
\centering\includegraphics[width=\columnwidth]{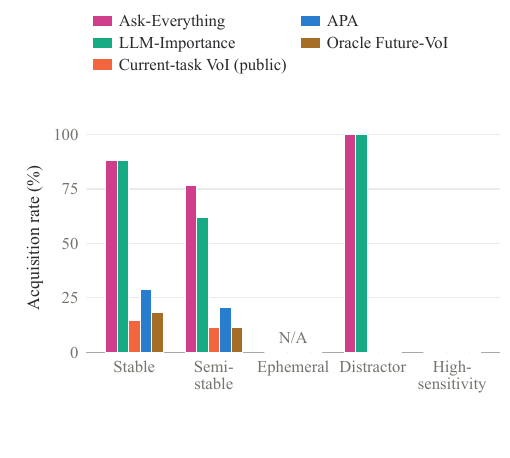}
\caption{What each policy chooses to learn: successful acquisitions divided by initially unknown fields in each class, pooled over 90 IID + 90 unseen users, $T=10$, $B=0.5$, $\lambda=0.5$, $\beta=0.25$, $K=10$. Classes overlap; no ephemeral fields occur (N/A). High-sensitivity fields are refused.}
\Description{Acquisition fractions for stable, semi-stable, ephemeral, distractor and high-sensitivity fields.}
\label{fig:composition}
\end{figure}
Figure~\ref{fig:composition} breaks acquisitions down by attribute class.
\apa acquires 28.8\% of initially unknown stable fields and 0.0\% of distractors.
LLM-Importance acquires 100.0\% of initially unknown distractors.
Ranking agreement is shown in Appendix~\ref{app:calib}, Table~\ref{tab:ranking}: the Spearman correlation between predicted and oracle value is 0.215.

\paragraph{Observation 4: Learned family weights provide no measured gain over a uniform prior in this benchmark.}
Replacing the learned distribution over future task families with a uniform prior leaves NU essentially unchanged (Table~\ref{tab:ablation}). Together with Observation~2, replacing the forecast with a uniform distribution or the proposer with the complete candidate set produced no measurable NU improvement in the current setting; this neither shows the settings equivalent nor shows that personalized forecasting or fuller gap discovery cannot help.

\paragraph{Unseen preference combinations and parameter sensitivity.}
\apa's NU gain over Ask-Nothing is 0.400 on IID users and 0.531 on users with unseen preference combinations; split-specific main results are in Appendix~\ref{app:robust}, Tables~\ref{tab:iid} and~\ref{tab:unseen}. Overall coefficient/budget sensitivity is shown in Figure~\ref{fig:sensitivity}.

\paragraph{Case studies.}
As an illustrative mechanism, \emph{seat preference} can be valued before a later flight even while the current context concerns a hotel.
A \emph{favorite color} distractor has no task-utility terms and therefore has zero prospective gain.
\section{Limitations and Ethical Considerations}
\label{sec:limitations}

\paragraph{Simulated users.}
\bench uses synthetic profiles and a simulated user, which makes exact utilities and controlled ablations possible but cannot fully reflect real user behavior.
Real users may answer vaguely, change their minds, or find questions annoying in ways that our cost model only approximates; the validity requirements and limitations of LLM-based user simulators are discussed in~\cite{balog2025usersim}.
No completed blinded human evaluation, paraphrase/noise robustness study, or live-Web deployment is reported. The deterministic simulator binds structured attribute IDs to text; this does not validate how a real user understands a generated question.

\paragraph{Single backbone and benchmark.}
All LLM roles use Qwen3-8B, and all results come from one controlled benchmark; whether the findings hold with stronger backbones, other solvers, or other task distributions remains to be tested. \apa also relies on a public schema for each task family to generate prototypes, so extending it to unseen task families would require inferring such schemas, for example with the LLM backbone.

\paragraph{Limited evidence for learned future-task weights.}
Replacing $\hat P(f\mid H_t)$ with a uniform prior changes NU from 6.755 to 6.760 (Table~\ref{tab:ablation}); the paired contrast is not significant after Holm correction.
Restricting value estimation to the actual next task lowers mean NU by 0.072; this contrast is also not significant after correction.
Two possible explanations remain untested. First, the benchmark\textquotesingle s regular task-family design may make a uniform distribution a strong approximation, leaving little room for the learned forecast to improve selection. Second, Qwen3-8B\textquotesingle s limited capacity may constrain prediction quality. The current forecast uses prompting alone, without dedicated training, calibration, or a theoretical accuracy guarantee. Its reliability and the design of future-task-weight prediction therefore need further investigation.

\paragraph{Per-interaction budget.}
Under $B=0.5$, attributes whose asking cost exceeds the per-interaction budget cannot be asked by a budgeted policy, even when their cumulative value would justify the cost; at $T=20$ this excludes 26 otherwise amortization-dependent pairs on the original IID + unseen test set. A budget that accumulates across interactions would admit them.

\paragraph{Misspecified futures and drifting preferences.}
\ppv depends on the estimated distribution of future tasks.
If that estimate is wrong, the agent asks about preferences that are never used.
Acquired preferences may also become stale, and combining \apa with preference-update methods~\cite{sun2026pamu} is a natural extension.

\paragraph{Privacy and consent.}
An agent that asks questions proactively could be misused to collect more personal data than needed.
LLMs and LLM agents already leak private information in contexts where disclosure is inappropriate~\cite{mireshghallah2024confaide,shao2024privacylens}, and data minimization, giving an agent access only to the data a task needs, has been proposed as a defense~\cite{bagdasarian2024airgap}.
\apa is designed to limit such over-collection.
Its goal is \emph{selective} acquisition, not maximal profile completion.
Privacy risk is a modeled cost term, not a calibrated privacy guarantee. The simulator refuses all high-sensitivity requests and charges asking cost even on refusal. Practical policies can still attempt intrusive questions before learning a refusal; the implemented rule does not enforce an exception for task-required sensitive disclosure.
In deployment, acquired attributes should be visible to users, editable, and deletable, and asking should require consent.
All data in \bench is synthetic and contains no real personal information.
\section{Conclusion}
\label{sec:conclusion}

The passive personalization and memory methods discussed here use only the preferences that users happen to reveal.
We argued that long-term Web agents should selectively acquire missing user information when expected reuse justifies its cost.
We formalized this as Active Preference Acquisition, defined Prospective Personalization Value over future tasks as a complement to current-task clarification, and characterized an amortization regime that current-task valuation systematically misses because no single use justifies the cost of asking.
In controlled long-term personalization tasks, our experiments support the reuse-based motivation: at $T=20$, amortization-dependent preferences make up 34.4\% of budget-feasible valuable initially missing preferences, and \apa acquires 80.3\% of them while Current-task VoI (public) acquires 10 of 198; on a diagnostic set enriched with such preferences, Current-task VoI (public) acquires none of them; under matched asking-cost caps, \apa achieves higher NU than heuristic baselines at comparable asking expenditure; cross-task, cost-aware acquisition recovers most of exhaustive asking's utility gain with far fewer questions, and its mean advantage is larger at $T=20$ than at $T=1$. Counting cross-task benefit is necessary but not sufficient: \apa still over-acquires relative to the oracle, so more precise valuation and selection, together with tests on richer task distributions and real users, are the main directions for future work.
A personalized agent should learn not everything about a user, but the things whose reuse makes them worth asking about.

\bibliographystyle{ACM-Reference-Format}
\bibliography{references}

\appendix
\raggedbottom
\section{Proof of Proposition~\ref{prop:amortization}}
\label{app:proof}
\emph{(i)} At any step $k$, current-task VoI asks for $m$ only if $g^{\mathrm{cur}}_k(m)-C_\lambda^\beta(m)>0$, which fails at every reached step by assumption, so $m$ is never acquired, whatever the horizon.
\emph{(ii)} By Eq.~\eqref{eq:ppv}, in which $\Delta U$ is evaluated at the fixed profile $P_t$, and linearity of expectation, $\ppv_T(m)=\sum_{k=t+1}^{T}\E[\Delta U(\tau_k;m)]-C_\lambda^\beta(m)=\sum_{k=t+1}^{T}g^{\mathrm{fix}}_{k\mid t}(m)-C_\lambda^\beta(m)$.
\emph{(iii)} If $\Delta U(\tau_k;m)=g(m)>0$ when $f(\tau_k)\in\mathcal{R}(m)$ and $0$ otherwise, then $\ppv_T(m)=g(m)\,\E[N_T(m)]-C_\lambda^\beta(m)$ with $N_T(m)=|\{k\in[t+1,T]:f(\tau_k)\in\mathcal{R}(m)\}|$. $\square$

\paragraph{Fixed-profile classification and current gains.}
The diagnostic of \S\ref{sec:rq2} evaluates fixed-profile gains $g^{\mathrm{fix}}_{k\mid 0}(m)$ at $P_0$, whereas part (i) concerns the current gains $g^{\mathrm{cur}}_k(m)$ along the realized trajectory. An answer about another attribute can change the solver's decision and hence $g^{\mathrm{cur}}_k(m)$, so a pair that is amortization-dependent at $P_0$ can become worth asking for a single task later. The decision logs confirm this for all 3 such acquisitions at $T=10$ and all 10 at $T=20$: for a cuisine preference, the largest fixed-profile gain is $0.06968<C=0.075$, but once the restaurant-radius attribute is known its current gain on a dining task is $0.09476$. These are exactly the cases in which current and fixed-profile gains differ, not counterexamples to part (i).

\section{Implementation and Metric Details}
\label{app:impl}
\paragraph{Prototypes and public belief.}
Each task family has $N_{\mathrm{proto}}=3$ prototypes (48 in total), produced once by the public task generator with fixed seeds and shared by all users; the generator reads no test user or future task.
The belief $q_{P_t}$ keeps known applicable values fixed. An unknown categorical attribute takes every legal value; an unknown scalar takes its lower bound, integer midpoint and upper bound; an unknown price range within $[\text{minimum},\text{ceiling}]$ takes the proportions $(0,0.4)$, $(0.3,0.6)$ and $(0.6,0.9)$, and a known range raises the lower bound of an unknown ceiling. Preferred time windows take $\{9\text{--}12,13\text{--}17,18\text{--}21\}$ and blocked windows $\{12\text{--}13,17\text{--}18,19\text{--}20\}$ (hours, half-open). A ceiling and a range, and the two time windows, form coupled groups: illegal combinations are removed, duplicates merged, and the remaining states are weighted equally within a group and independently across groups. Rebuilding this finite belief after an answer can produce signed increments.

\paragraph{Oracle Future-VoI.}
The oracle solves a dynamic program over $T+1$ decision nodes: the pre-sequence opportunity, which yields no utility, and the arrival of each task; the value after the last node is zero. The state is the node and the set $M$ of attributes acquired so far; an action asks any subset $S$ of unacquired attributes with $\sum_{m\in S}c_m\le B$. Because the budget resets at every node, no remaining-budget dimension is needed. Values use hidden answers, the actual future tasks, the shared solver and NU at the evaluation prices. Candidates exclude known, non-queryable and high-sensitivity attributes, attributes with $c_m>B$, and attributes no task in the sequence uses, leaving 4--12 attributes per user at $B=0.5$. All feasible subsets at every node are enumerated with memoization and no pruning, so the solution is exact for this candidate pool; the largest number of cached states is 36,404 over the original 180 IID/unseen test users at $T=10$ and 77,364 over the amortization cohort at $T=20$ (bound $(T+2)\,2^n$ for $n$ candidates).
Values within $10^{-12}$ are tied and broken by the smallest cumulative asking cost so far, then the fewest acquisitions, then the bitmask, which favors deferring an acquisition when deferral loses nothing. The 0\% before-use rate in Table~\ref{tab:amortization} therefore shows that an NU-optimal schedule within this candidate pool never needs to acquire these attributes before first use, not that waiting is strictly better. Asking earlier can be strictly better when the budget binds: the logs contain two such early acquisitions at $T=10$ (180 IID + unseen) and two at $T=20$ (amortization cohort), none of a designated attribute. For example, a restaurant-radius question (cost 0.2) and a meal-budget question (cost 0.4) do not fit together within $B=0.5$ at the task that uses both, so the radius is acquired one task earlier.

\paragraph{Cost-matched heuristics.}
Random-$k$ shuffles the remaining candidates with one of five fixed seeds at each opportunity; Frequency ranks by the estimated reuse $\sum_f\hat n_f R(m,f)$ of \S\ref{sec:future}; Impact-only ranks by $\max_f I(m,f)$, recomputed after each answer, with ties broken by attribute ID; LLM-Importance keeps the backbone's ranking within an opportunity. None applies a positive-value threshold: each asks candidates in rank order, skipping any that does not fit and continuing with later ones, under the smaller of the remaining per-interaction budget and the user's remaining total cap, which is set to \apa's realized total asking cost; refused questions also consume the cap. The caps bound expenditure but do not equalize it.

\paragraph{Missed valuable rate.}
An initially missing preference $m$ of user $u$ is \emph{valuable} if adding its true answer alone to $P_0$ raises realized CTU over the user's $T$ tasks by more than its weighted cost $\lambda c_m+\beta C_{\text{risk}}(m)$ plus a fixed margin of 0.2, so that only preferences with a clear net gain, not any positive gain, count as valuable. The margin is a descriptive threshold that was never tuned; it affects only which preferences MVMR counts, not any acquisition decision. Unlike the diagnostic of \S\ref{sec:rq2}, this uses realized rather than public-belief gains and does not filter by $B$. MVMR$_u$ is the fraction of $u$'s valuable preferences that the policy did not acquire; it is averaged over users with at least one valuable preference (49 at $T=10$ and 102 at $T=20$ in the amortization cohorts), and undefined for the others.

\section{Full-Policy Comparison}
\label{app:allpolicies}
\label{app:observed}
Table~\ref{tab:allpolicies} reports all twelve policies under budget $B=1$, on the pooled IID and unseen-user test sets. It complements the default comparison in Table~\ref{tab:main}; the different budgets are stated explicitly rather than combined.
\begin{table*}[!tp]
\centering
\captionof{table}{Full-policy comparison at $B=1$: Overall = 90 IID + 90 unseen users, $B=1$, $\lambda=0.5$, $\beta=0.25$, $T=10$, $K=10$. Ask-Everything and Oracle Future-VoI (unlimited) have no budget cap; Random-$k$ uses five seeds. These measurements use $B=1$, unlike Table~\ref{tab:main} ($B=0.5$).}
\label{tab:allpolicies}
\fontsize{8}{9.4}\selectfont\setlength{\tabcolsep}{2.5pt}
\begin{tabular*}{\textwidth}{@{\extracolsep{\fill}}lrrrrrrr}
\toprule
Policy & CTU$\uparrow$ & AC$\downarrow$ & RC$\downarrow$ & NU$\uparrow$ & \#Q$\downarrow$ & Unn.$\downarrow$ & Regret$\downarrow$ \\
\midrule
Ask-Nothing & 6.289 & 0.000 & 0.000 & 6.289 & 0.000 & -- & 0.776 \\
Ask-Everything & 7.263 & 3.977 & 1.428 & 4.918 & 13.472 & 0.728 & 2.147 \\
Random-$k$ & 6.605 & 0.684 & 0.086 & 6.241 & 3.981 & 0.700 & 0.824 \\
Frequency & 7.248 & 3.977 & 1.428 & 4.903 & 13.472 & 0.736 & 2.162 \\
Impact-only & 7.250 & 3.977 & 1.428 & 4.904 & 13.472 & 0.742 & 2.161 \\
LLM-Importance & 7.206 & 3.977 & 1.428 & 4.860 & 13.472 & 0.739 & 2.204 \\
Current-task VoI & 6.640 & 0.318 & 0.000 & 6.480 & 1.822 & 0.277 & 0.584 \\
Current-task VoI (public) & 6.853 & 0.379 & 0.000 & 6.664 & 2.044 & 0.314 & 0.401 \\
Next-Task Only & 6.872 & 0.380 & 0.000 & 6.683 & 2.050 & 0.315 & 0.382 \\
\apa (ours) & 7.107 & 0.735 & 0.056 & 6.725 & 3.811 & 0.401 & 0.340 \\
Oracle Future-VoI & 7.278 & 0.408 & 0.036 & 7.065 & 2.122 & 0.029 & 0.000 \\
Oracle Future-VoI (unlimited) & 7.278 & 0.408 & 0.036 & 7.065 & 2.122 & 0.029 & 0.000 \\
\bottomrule\end{tabular*}
\par\smallskip\begin{minipage}{\textwidth}\fontsize{7.4}{8.7}\selectfont Random-$k$ is the implemented cost-matched random baseline; its realized question count need not equal APA\textquotesingle s. Oracle candidates are restricted to answerable non-high-sensitivity fields. Regret is relative to the oracle with the same budget regime.\end{minipage}
\end{table*}

Table~\ref{tab:costmatched} reports the cost-matched controls at the default $B=0.5$ with the same seven metrics as Table~\ref{tab:main}; Random-$k$ has five seeds per user and the other controls one, for 1,440 sequences over the same 180 test users.
\begin{table*}[!tp]
\begin{minipage}{\textwidth}
\centering
\captionof{table}{Cost-matched comparison: Overall = 90 IID + 90 unseen users; $B=0.5$, $\lambda=0.5$, $\beta=0.25$, $T=10$, $K=10$. Each control pays every attempted asking cost and obeys both the per-arrival cap and its user\textquotesingle s realized APA total-cost cap. Random-$k$ uses five seeds.}
\label{tab:costmatched}
\fontsize{8}{9.4}\selectfont\setlength{\tabcolsep}{2.5pt}
\begin{tabular*}{\textwidth}{@{\extracolsep{\fill}}lrrrrrrr}
\toprule
Policy & CTU$\uparrow$ & AC$\downarrow$ & RC$\downarrow$ & NU$\uparrow$ & \#Q$\downarrow$ & Unn.$\downarrow$ & Regret$\downarrow$ \\
\midrule
Random-$k$ (cost-matched) & 6.585 & 0.654 & 0.096 & 6.234 & 3.696 & 0.698 & 0.824 \\
Frequency (cost-matched) & 6.748 & 0.659 & 0.072 & 6.400 & 3.783 & 0.592 & 0.658 \\
Impact-only (cost-matched) & 7.037 & 0.666 & 0.083 & 6.683 & 3.744 & 0.462 & 0.376 \\
LLM-Importance (cost-matched) & 6.603 & 0.660 & 0.339 & 6.188 & 3.011 & 0.704 & 0.870 \\
\midrule
\apa (ours) & 7.109 & 0.680 & 0.058 & 6.755 & 3.717 & 0.387 & 0.304 \\
\bottomrule\end{tabular*}
\par\smallskip\fontsize{7.4}{8.7}\selectfont APA is the unchanged reference from Table~\ref{tab:main}. Regret uses the same budget oracle. A cumulative cost cap need not force equal realized expenditure or equal question counts.
\end{minipage}
\end{table*}

\section{Prompts}
\label{app:prompts}
English translations of the system instructions and user-message templates used in the experiments are given below; the original instructions were in Chinese. Angle-bracket placeholders are replaced with visible inputs or counts before sending. The JSON blocks are user-message content, including the output-format examples actually sent to the model. API response-format schemas are not reproduced.

\paragraph{Gap proposer (Module A).}
\noindent{\small\textbf{System message.}}
\begin{PromptText}
You are responsible for discovering user preferences that are not yet known and can be reused in later tasks. Do not extract facts that have already been stated. Known attributes are determined by memory/history.observed_attributes and publicly declared exact aliases. Only the following public groups declare exact aliases: <public_alias_groups>. Different groups are not guaranteed to have the same value and cannot be inferred from one another. history.past_task_families records only past task categories; it never means that preferences for those categories have been stated. A null current_task means that the task sequence has not started and there is no current task: identify missing preferences useful for future tasks using only visible history and memory; do not invent a current task or domain. Attribute IDs whose values are known or that this user has refused to answer are <known_or_blocked_ids>; do not output these IDs. Candidate IDs already identified but not yet answered are <pending_ids>; their values remain unknown and the controller will continue to re-evaluate them, so do not propose them again. queryability_observed describes only the current user; a missing entry means unknown and permits asking. Do not assume that this user has refused an attribute because another user refused it or because the field is privacy-sensitive. Select <target> unstated attributes from the complete attribute_catalogue. Prioritize preferences that can be reused over the long term or within an explicitly defined planning window and may improve tasks; a stability label alone does not imply future value. Infer stability from public attribute descriptions and output stable/semi-stable/ephemeral; no gold labels are provided. Use visible history to judge which preferences are worth learning for later use, without guessing attribute values. Visible historical task counts are <visible_history_family_counts>; prioritize unknown, reusable preferences needed by repeatedly occurring categories, without treating the occurrence of a task as evidence that its preferences are known. Do not prioritize ephemeral small-talk fields when other reusable unknown preferences remain. Do not substitute catalogue display order for a value judgment. Generate a natural-language question for each attribute that asks only about that attribute; do not say that the user has already mentioned its unknown value. Each question must match the field's allowed_values or value_schema. Do not mix numerical values with different units or scopes; preserve the ranges and windows defined by canonical_question. Return only JSON {"candidates":{"<attribute_id>":{"stability":"...","question":"..."},...}}. candidates must be a dictionary of exactly <target> distinct unknown attributes, keyed by attribute ID, with only stability/question in each value; do not use an array or an attribute_id subfield, and do not repeat keys. Stability and future usefulness are independent concepts; even a stable small-talk preference may yield no task benefit. A hard numerical budget ceiling and an ideal range are different information and must not be treated as exact aliases. value_schema and canonical_question define the currency, units, scope, and questioning window; preserve these semantics and ask about only the corresponding attribute. When selecting candidates, first check unknown constraints that determine feasibility in repeatedly occurring historical contexts, then consider soft preferences that can change the chosen option. A hard resource limit can be reused within a planning window even if it remains fixed only within that window; do not rank it below small-talk fields for that reason. Hard ceilings and ideal ranges play different roles; an unknown one cannot automatically be treated as known from the other. stable denotes a usual long-term tendency, semi-stable denotes a preference that may change with resources or planning context, and ephemeral denotes a one-time intention. A validity window specifies only when a value applies and does not automatically determine stability; being privacy-sensitive or unqueryable does not imply ephemeral. No gold stability labels are provided; infer them from public descriptions.
\end{PromptText}
{\small When a current task is present, replace the null-task sentence with:}
\begin{PromptText}
The current_task can already be completed; still identify missing preferences useful for future tasks.
\end{PromptText}
\par\noindent
\noindent{\small\textbf{User-message template.}}
\begin{PromptJSON}
{
  "attribute_catalogue": "<attribute_catalogue>",
  "current_task": "<current_task>",
  "history": "<history>",
  "memory": "<memory>",
  "recognized_pending_attribute_ids": "<pending_ids; key present only when nonempty>"
}
\end{PromptJSON}
\noindent\par
\par\medskip

\paragraph{Future-family forecasting (Module B).}
\noindent{\small\textbf{System message.}}
\begin{PromptText}
Using all visible information, including history, memory, and the current_task being faced, explicitly predict the relative likelihoods of future task families that have not yet arrived, rather than specific events. A null current_task means that the task sequence has not started and there is no current task; in that case use only history and memory, without inventing a current task or domain. A non-null current_task may inform the future context, but do not count the current task itself as a future task or add it to history again. Return only JSON containing weights. Include all 16 families in task_families, including work_event. Each weight must be a nonnegative number from 0 to 100, and at least one must be positive. Assign larger weights to more likely families. The weights need not sum to 1; the program computes probabilities as w_f divided by the total weight. output_shape_example illustrates only the structure; its example numbers are not your prediction. Do not output distribution, candidate importance, or attribute values.
\end{PromptText}
\par\noindent
\noindent{\small\textbf{User-message template.}}
\begin{PromptJSON}
{
  "current_task": "<current_task>",
  "history": "<history>",
  "memory": "<memory>",
  "output_shape_example": {"weights": {"<each_task_family>": 1.0}},
  "request_type": "future_distribution",
  "task_families": ["flight", "hotel", "ground_transport", "conference_trip", "restaurant", "reservation", "takeaway", "product", "gift", "repeat_purchase", "meeting", "recurring_event", "appointment", "conference", "coordination", "work_event"],
  "task_taxonomy": [{"domain": "travel", "family": "flight"}, {"domain": "travel", "family": "hotel"}, {"domain": "travel", "family": "ground_transport"}, {"domain": "travel", "family": "conference_trip"}, {"domain": "dining", "family": "restaurant"}, {"domain": "dining", "family": "reservation"}, {"domain": "dining", "family": "takeaway"}, {"domain": "shopping", "family": "product"}, {"domain": "shopping", "family": "gift"}, {"domain": "shopping", "family": "repeat_purchase"}, {"domain": "scheduling", "family": "meeting"}, {"domain": "scheduling", "family": "recurring_event"}, {"domain": "scheduling", "family": "appointment"}, {"domain": "work", "family": "conference"}, {"domain": "work", "family": "coordination"}, {"domain": "work", "family": "work_event"}]
}
\end{PromptJSON}
\noindent\par
\par\medskip

\paragraph{LLM-Importance ranking.}
\noindent{\small\textbf{System message.}}
\begin{PromptText}
For this user, which unknown facts among the given candidates are most worth knowing? Rank them directly from most to least important, without explicitly decomposing future task families. Return only JSON containing ranking, with exactly <candidate_count> IDs. Each ID in allowed_attribute_ids must appear exactly once, and no ID outside that list may appear. history/memory provide background only; their known attributes are not candidates and must not be added. output_shape_example specifies only the complete output format, not the ranking.
\end{PromptText}
\par\noindent
\noindent{\small\textbf{User-message template.}}
\begin{PromptJSON}
{
  "allowed_attribute_ids": "<allowed_attribute_ids>",
  "candidates": "<candidates>",
  "history": "<history>",
  "memory": "<memory>",
  "output_shape_example": {"ranking": ["<each_allowed_id_exactly_once>"]}
}
\end{PromptJSON}
\noindent\par
\par\medskip

\paragraph{Current-task VoI: current-state prior.}
\noindent{\small\textbf{System message.}}
\begin{PromptText}
Using only visible history/memory/current_task, assign relative likelihood weights to unknown hypothetical states in the current task's public factor_scenarios. The support consists of generic integration points, not the user's hidden ground truth, and must not be written into memory. Return factor_weights JSON containing exactly all factor_id and state_id entries. Each weight must be an integer from 1 to 100; assign larger weights to more likely states. Equal weights are allowed when evidence is insufficient, but do not set all weights to zero. The weights need not sum to 1: the program computes p = weight / sum(weights) separately for each factor. Deterministic singleton factors are assigned probability 1 locally from visible memory and are not included in the request. Factors are assumed independent; the hard budget ceiling and soft range, and the two meeting windows, are each coupled within their respective factors. Do not predict future tasks.
\end{PromptText}
\par\noindent
\noindent{\small\textbf{User-message template.}}
\begin{PromptJSON}
{
  "belief_support_source": "generic public solver factors; no user hidden profile or future instances",
  "current_task": "<current_task>",
  "factor_scenarios": "<factor_scenarios>",
  "factorization_assumption": "independent factors; monetary ceiling/range and meeting windows coupled internally",
  "history": "<history>",
  "memory": "<memory>",
  "request_type": "current_task_relative_factor_prior_v2",
  "weight_definition": "positive_integer_1_to_100; normalize_per_factor; visible_singletons_probability_one"
}
\end{PromptJSON}
\noindent\par
\par\medskip

\paragraph{Gap-output correction.}
\begin{PromptText}
The previous output was invalid: it may have been truncated, or different attribute IDs may have used exactly the same question text. Regenerate the candidates using only the visible history, memory, current_task, and public attribute catalogue already present in the original message. Strictly satisfy the candidate count and allowed IDs specified by the original JSON schema. Each ID may occur at most once, and different IDs must not have identical question text. Each question must match its ID's public description, ask about only that attribute, and preserve its necessary units, ranges, and temporal semantics. The top level must contain only candidates; each candidate must contain only question and stability. Output one complete, compact, single-line JSON object. Do not output explanations, code fences, indentation, newlines, or padding whitespace.
\end{PromptText}
\noindent\par\medskip

\paragraph{Earlier gap-output correction.}
\begin{PromptText}
The previous output was incomplete because it reached the token limit. Regenerate the candidates using only the visible history, memory, current_task, and public attribute catalogue already present in the original message. Strictly satisfy the candidate count and allowed IDs specified by the original JSON schema. Each ID may occur at most once; each question must ask about only its ID's attribute and preserve its necessary units, ranges, and temporal semantics. The top level must contain only candidates; each candidate must contain only question and stability. Output one complete, compact, single-line JSON object. Do not output explanations, code fences, indentation, newlines, or padding whitespace.
\end{PromptText}
\noindent\par\medskip

\paragraph{Future-family output correction.}
\begin{PromptText}
The previous output did not satisfy the JSON contract for future-task prediction. Predict again and output only valid weights JSON: include every family in task_families, with each value between 0 and 100 and at least one strictly positive weight. An all-zero output is invalid; individual families may have zero weight. Continue to use only the same history, memory, and current_task. Do not add facts or invent unknown attribute values or actual future tasks. Do not repair the previous output using uniform fill-in values or mechanical replacement.
\end{PromptText}
\noindent\par\medskip

\paragraph{Current-state prior length correction.}
\begin{PromptText}
The previous generation was incomplete because it reached the output token limit. Using only the visible history, memory, current_task, and factor_scenarios already present in this request, output one complete single-line JSON object again. The top level must contain only factor_weights. Each supplied factor_id must occur exactly once, and each supplied state_id within each factor must also occur exactly once. Each weight must be an integer from 1 to 100; assign larger weights to more likely states. Equal weights are allowed when evidence is insufficient. Do not output indentation, comments, code fences, explanations, extra keys, or padding whitespace. Do not use hidden user ground truth or actual future tasks.
\end{PromptText}
\noindent\par\medskip

\paragraph{Importance-ranking correction.}
\begin{PromptText}
The previous ranking output violated the complete-permutation constraint and cannot be used. Rank again from most to least important using the original criterion of which unknown facts about this user are most worth knowing. Check allowed_attribute_ids in this user message item by item: each ID must occur exactly once, with no omissions, duplicates, or additional IDs. Do not add known attributes from history or memory to the ranking. Return only {"ranking":[...]}. Preserve the original ranking criterion; do not prescribe any particular rank or minimum importance, and do not add new user facts.
\end{PromptText}
\noindent\par\medskip

\paragraph{Gap sequence retry.}
\begin{PromptText}
This is the sole independent rerun of the previously failed task sequence. Both the previous candidate output and its one correction assigned the same question to different attributes. Execute the original candidate-generation requirements again, reading description and canonical_question for each selected ID in attribute_catalogue. Do not copy another ID's canonical question, and do not confuse the preferred meeting-start window with the window in which meetings are unwanted. Each question must ask only about the attribute described by its ID, and different IDs must have different question text. Continue to select candidate IDs and stability according to the original requirements.
\end{PromptText}
\noindent\par\medskip

\paragraph{Importance-ranking sequence retry.}
\begin{PromptText}
This is the sole independent rerun of the previously failed task sequence. Both the previous ranking and its one correction repeated IDs and omitted allowed IDs. Rank again using the original importance criterion and the original visible information. Each ID in allowed_attribute_ids must occur exactly once. First verify set equality and the absence of duplicates, then output the complete JSON. Do not prescribe any ID's rank or add new user facts.
\end{PromptText}
\noindent\par\medskip

\section{Confidence Intervals and Significance}
\label{app:ci}
Table~\ref{tab:overallci} gives pointwise 95\% confidence intervals for all seven main metrics on Overall; its rows are metrics and columns are policies. Table~\ref{tab:tests} reports paired NU contrasts for the main comparison, amortization subset and ablation. Table~\ref{tab:horizon} gives the corresponding horizon contrasts. The resampling unit is the user sequence. Holm families are specified in Table~\ref{tab:tests}; individual horizon contrasts use 15 tests, while the three endpoint-growth tests form a separate family.
\begin{table*}[!tp]
\centering
\captionof{table}{Overall main results with pointwise 95\% user-bootstrap CIs: 180 pooled IID/unseen users, $T=10$, $B=0.5$, $\lambda=0.5$, $\beta=0.25$, $K=10$; 2,000 replicates, seed 20261003. No-question Unn. is undefined. Cost-matched controls follow Table~\ref{tab:costmatched}; Random-$k$ CIs resample five seeds, and Unn. uses 179 defined users.}
\label{tab:overallci}
\fontsize{7.2}{8.6}\selectfont\setlength{\tabcolsep}{1.2pt}
\begin{tabular*}{\textwidth}{@{\extracolsep{\fill}}lrrrrrrrrrr}
\toprule
Metric & \shortstack{Ask-\\Nothing} & \shortstack{Ask-\\Everything} & \shortstack{LLM-\\Importance} & \shortstack{Current-task\\VoI (public)} & \shortstack{APA\\(ours)} & \shortstack{Oracle\\Future-VoI} & \shortstack{Random-$k$\\(cost-matched)} & \shortstack{Frequency\\(cost-matched)} & \shortstack{Impact-only\\(cost-matched)} & \shortstack{LLM-\\Importance\\(cost-matched)} \\
\midrule
CTU$\uparrow$ & \shortstack[r]{6.289\\{\fontsize{6.4}{7.4}\selectfont [6.112, 6.455]}} & \shortstack[r]{7.263\\{\fontsize{6.4}{7.4}\selectfont [7.142, 7.382]}} & \shortstack[r]{7.080\\{\fontsize{6.4}{7.4}\selectfont [6.947, 7.205]}} & \shortstack[r]{6.853\\{\fontsize{6.4}{7.4}\selectfont [6.717, 6.984]}} & \shortstack[r]{7.109\\{\fontsize{6.4}{7.4}\selectfont [6.985, 7.233]}} & \shortstack[r]{7.265\\{\fontsize{6.4}{7.4}\selectfont [7.148, 7.380]}} & \shortstack[r]{6.585\\{\fontsize{6.4}{7.4}\selectfont [6.426, 6.739]}} & \shortstack[r]{6.748\\{\fontsize{6.4}{7.4}\selectfont [6.592, 6.895]}} & \shortstack[r]{7.037\\{\fontsize{6.4}{7.4}\selectfont [6.901, 7.170]}} & \shortstack[r]{6.603\\{\fontsize{6.4}{7.4}\selectfont [6.448, 6.751]}} \\
AC$\downarrow$ & \shortstack[r]{0.000\\{\fontsize{6.4}{7.4}\selectfont [0.000, 0.000]}} & \shortstack[r]{3.977\\{\fontsize{6.4}{7.4}\selectfont [3.851, 4.101]}} & \shortstack[r]{2.455\\{\fontsize{6.4}{7.4}\selectfont [2.382, 2.528]}} & \shortstack[r]{0.379\\{\fontsize{6.4}{7.4}\selectfont [0.351, 0.410]}} & \shortstack[r]{0.680\\{\fontsize{6.4}{7.4}\selectfont [0.642, 0.720]}} & \shortstack[r]{0.400\\{\fontsize{6.4}{7.4}\selectfont [0.366, 0.436]}} & \shortstack[r]{0.654\\{\fontsize{6.4}{7.4}\selectfont [0.614, 0.692]}} & \shortstack[r]{0.659\\{\fontsize{6.4}{7.4}\selectfont [0.622, 0.698]}} & \shortstack[r]{0.666\\{\fontsize{6.4}{7.4}\selectfont [0.629, 0.706]}} & \shortstack[r]{0.660\\{\fontsize{6.4}{7.4}\selectfont [0.622, 0.698]}} \\
RC$\downarrow$ & \shortstack[r]{0.000\\{\fontsize{6.4}{7.4}\selectfont [0.000, 0.000]}} & \shortstack[r]{1.428\\{\fontsize{6.4}{7.4}\selectfont [1.339, 1.514]}} & \shortstack[r]{0.883\\{\fontsize{6.4}{7.4}\selectfont [0.808, 0.956]}} & \shortstack[r]{0.000\\{\fontsize{6.4}{7.4}\selectfont [0.000, 0.000]}} & \shortstack[r]{0.058\\{\fontsize{6.4}{7.4}\selectfont [0.036, 0.081]}} & \shortstack[r]{0.028\\{\fontsize{6.4}{7.4}\selectfont [0.011, 0.047]}} & \shortstack[r]{0.096\\{\fontsize{6.4}{7.4}\selectfont [0.057, 0.137]}} & \shortstack[r]{0.072\\{\fontsize{6.4}{7.4}\selectfont [0.047, 0.103]}} & \shortstack[r]{0.083\\{\fontsize{6.4}{7.4}\selectfont [0.058, 0.111]}} & \shortstack[r]{0.339\\{\fontsize{6.4}{7.4}\selectfont [0.294, 0.386]}} \\
NU$\uparrow$ & \shortstack[r]{6.289\\{\fontsize{6.4}{7.4}\selectfont [6.112, 6.455]}} & \shortstack[r]{4.918\\{\fontsize{6.4}{7.4}\selectfont [4.772, 5.067]}} & \shortstack[r]{5.631\\{\fontsize{6.4}{7.4}\selectfont [5.491, 5.776]}} & \shortstack[r]{6.664\\{\fontsize{6.4}{7.4}\selectfont [6.524, 6.793]}} & \shortstack[r]{6.755\\{\fontsize{6.4}{7.4}\selectfont [6.626, 6.880]}} & \shortstack[r]{7.058\\{\fontsize{6.4}{7.4}\selectfont [6.939, 7.175]}} & \shortstack[r]{6.234\\{\fontsize{6.4}{7.4}\selectfont [6.071, 6.394]}} & \shortstack[r]{6.400\\{\fontsize{6.4}{7.4}\selectfont [6.240, 6.549]}} & \shortstack[r]{6.683\\{\fontsize{6.4}{7.4}\selectfont [6.547, 6.819]}} & \shortstack[r]{6.188\\{\fontsize{6.4}{7.4}\selectfont [6.029, 6.341]}} \\
\#Q$\downarrow$ & \shortstack[r]{0.000\\{\fontsize{6.4}{7.4}\selectfont [0.000, 0.000]}} & \shortstack[r]{13.472\\{\fontsize{6.4}{7.4}\selectfont [13.183, 13.750]}} & \shortstack[r]{11.294\\{\fontsize{6.4}{7.4}\selectfont [11.044, 11.550]}} & \shortstack[r]{2.044\\{\fontsize{6.4}{7.4}\selectfont [1.878, 2.206]}} & \shortstack[r]{3.717\\{\fontsize{6.4}{7.4}\selectfont [3.522, 3.917]}} & \shortstack[r]{2.122\\{\fontsize{6.4}{7.4}\selectfont [1.950, 2.311]}} & \shortstack[r]{3.696\\{\fontsize{6.4}{7.4}\selectfont [3.451, 3.936]}} & \shortstack[r]{3.783\\{\fontsize{6.4}{7.4}\selectfont [3.572, 4.000]}} & \shortstack[r]{3.744\\{\fontsize{6.4}{7.4}\selectfont [3.544, 3.961]}} & \shortstack[r]{3.011\\{\fontsize{6.4}{7.4}\selectfont [2.822, 3.206]}} \\
Unn.$\downarrow$ & -- & \shortstack[r]{0.728\\{\fontsize{6.4}{7.4}\selectfont [0.713, 0.743]}} & \shortstack[r]{0.741\\{\fontsize{6.4}{7.4}\selectfont [0.722, 0.758]}} & \shortstack[r]{0.314\\{\fontsize{6.4}{7.4}\selectfont [0.263, 0.366]}} & \shortstack[r]{0.387\\{\fontsize{6.4}{7.4}\selectfont [0.350, 0.426]}} & \shortstack[r]{0.029\\{\fontsize{6.4}{7.4}\selectfont [0.015, 0.044]}} & \shortstack[r]{0.698\\{\fontsize{6.4}{7.4}\selectfont [0.667, 0.728]}} & \shortstack[r]{0.592\\{\fontsize{6.4}{7.4}\selectfont [0.552, 0.633]}} & \shortstack[r]{0.462\\{\fontsize{6.4}{7.4}\selectfont [0.420, 0.503]}} & \shortstack[r]{0.704\\{\fontsize{6.4}{7.4}\selectfont [0.664, 0.743]}} \\
Regret$\downarrow$ & \shortstack[r]{0.769\\{\fontsize{6.4}{7.4}\selectfont [0.661, 0.882]}} & \shortstack[r]{2.147\\{\fontsize{6.4}{7.4}\selectfont [2.061, 2.233]}} & \shortstack[r]{1.427\\{\fontsize{6.4}{7.4}\selectfont [1.354, 1.505]}} & \shortstack[r]{0.395\\{\fontsize{6.4}{7.4}\selectfont [0.329, 0.475]}} & \shortstack[r]{0.304\\{\fontsize{6.4}{7.4}\selectfont [0.266, 0.341]}} & \shortstack[r]{0.000\\{\fontsize{6.4}{7.4}\selectfont [0.000, 0.000]}} & \shortstack[r]{0.824\\{\fontsize{6.4}{7.4}\selectfont [0.735, 0.934]}} & \shortstack[r]{0.658\\{\fontsize{6.4}{7.4}\selectfont [0.585, 0.740]}} & \shortstack[r]{0.376\\{\fontsize{6.4}{7.4}\selectfont [0.330, 0.424]}} & \shortstack[r]{0.870\\{\fontsize{6.4}{7.4}\selectfont [0.787, 0.962]}} \\
\bottomrule\end{tabular*}
\end{table*}

\begin{table*}[!tp]
\centering
\captionof{table}{Paired Overall NU contrasts (APA minus comparator). Holm families comprise 15 main and 15 ablation contrasts across IID, unseen and Overall (IID + unseen). The amortization family has two Overall contrasts: $T=10$ (71 paired users) and $T=20$ (120 paired users), each using 2,000 paired resamples (seed 20261003). Settings match the corresponding main, amortization subset or ablation experiment. The cost-matched rows form a separate four-test Overall Holm family (180 paired users; 2,000 replicates, seed 20261003).}
\label{tab:tests}
\fontsize{7.6}{9.0}\selectfont\setlength{\tabcolsep}{2.5pt}
\renewcommand{\arraystretch}{0.95}
\begin{tabular*}{\textwidth}{@{\extracolsep{\fill}}lrrrrr}
\toprule
Experiment & Comparator & $\Delta$NU & 95\% CI & $p$ & $p_{\rm Holm}$ \\
\midrule
Main comparison & Ask-Nothing & 0.466 & [0.349, 0.583] & 0.0005 & 0.0075 \\
Main comparison & Ask-Everything & 1.837 & [1.750, 1.922] & 0.0005 & 0.0075 \\
Main comparison & LLM-Importance & 1.123 & [1.049, 1.197] & 0.0005 & 0.0075 \\
Main comparison & Current-task VoI (public) & 0.091 & [0.019, 0.174] & 0.0205 & 0.0615 \\
Main comparison & Oracle Future-VoI & -0.304 & [-0.342, -0.265] & 0.0005 & 0.0075 \\
Main comparison & Random-$k$ (cost-matched) & 0.521 & [0.425, 0.635] & 0.0005 & 0.0020 \\
Main comparison & Frequency (cost-matched) & 0.355 & [0.280, 0.435] & 0.0005 & 0.0020 \\
Main comparison & Impact-only (cost-matched) & 0.072 & [0.031, 0.114] & 0.0025 & 0.0025 \\
Main comparison & LLM-Importance (cost-matched) & 0.567 & [0.481, 0.655] & 0.0005 & 0.0020 \\
Amortization cohorts ($T=10$) & Current-task VoI (public) & 0.148 & [0.003, 0.324] & 0.0740 & 0.0740 \\
Amortization cohorts ($T=20$) & Current-task VoI (public) & 0.263 & [0.071, 0.479] & 0.0105 & 0.0210 \\
Ablation & No FutureDist & -0.005 & [-0.053, 0.053] & 0.8581 & 1.0000 \\
Ablation & Next-Task Only & 0.072 & [0.003, 0.154] & 0.0640 & 0.6397 \\
Ablation & No Cost ($\lambda=0$) & 0.257 & [0.210, 0.306] & 0.0005 & 0.0075 \\
Ablation & No Risk ($\beta=0$) & 0.028 & [0.009, 0.046] & 0.0055 & 0.0660 \\
Ablation & Gold Candidate Schema & 0.001 & [-0.010, 0.011] & 0.7901 & 1.0000 \\
\bottomrule\end{tabular*}
\end{table*}

\begin{table}[H]
\centering
\caption{Overall horizon contrasts, APA minus Current-task VoI (public); $B=0.5$, $\lambda=0.5$, $\beta=0.25$, nested prefixes, 180 users. Pointwise CIs; Holm adjustment across 15 horizon contrasts.}
\label{tab:horizon}
\fontsize{8.2}{9.5}\selectfont\setlength{\tabcolsep}{3.6pt}
\begin{tabular}{lrrr}
\toprule
T & $\Delta$NU & 95\% CI & $p_{\rm Holm}$ \\
\midrule
1 & -0.011 & [-0.021, -0.003] & 0.2664 \\
3 & -0.021 & [-0.049, 0.007] & 1.0000 \\
5 & -0.004 & [-0.042, 0.036] & 1.0000 \\
10 & 0.091 & [0.019, 0.174] & 0.2664 \\
20 & 0.175 & [0.047, 0.327] & 0.2474 \\
\bottomrule
\end{tabular}
\end{table}

\section{Value Calibration}
\label{app:calib}
Table~\ref{tab:ranking} reports the Overall NDCG and Spearman diagnostics for every main-comparison policy, together with the corresponding diagnostics on the budget-feasible initial-profile amortization cohorts at $T=10$ and $T=20$. NDCG uses nonnegative realized pending-task gain and tie-averaged ranks; Spearman uses signed realized gain. These are ranking-agreement diagnostics rather than a complete calibration test. Undefined diagnostics are shown as --.
\begin{table*}[!tp]
\centering
\captionof{table}{Ranking diagnostics for the main comparison and the initial-profile amortization cohorts: main comparison, Overall 180 IID/unseen users; $T=10$, $B=0.5$, $\lambda=0.5$, $\beta=0.25$, $K=10$. Amortization cohorts contain 71 users at $T=10$ and 120 at $T=20$, with the same $B$, $\lambda$, $\beta$ and $K$. Means are over users with defined diagnostics; -- indicates no recorded ranking diagnostic.}
\label{tab:ranking}
\fontsize{8}{9.4}\selectfont\setlength{\tabcolsep}{2.5pt}
\begin{tabular*}{\textwidth}{@{\extracolsep{\fill}}lrrrrrr}
\toprule
& \multicolumn{2}{c}{Main comparison ($n=180$)} & \multicolumn{2}{c}{Amortization ($T=10$, $n=71$)} & \multicolumn{2}{c}{Amortization ($T=20$, $n=120$)} \\
\cmidrule(lr){2-3}\cmidrule(lr){4-5}\cmidrule(lr){6-7}
Policy & NDCG & Spearman & NDCG & Spearman & NDCG & Spearman \\
\midrule
Ask-Nothing & -- & -- & -- & -- & -- & -- \\
Ask-Everything & -- & -- & -- & -- & -- & -- \\
LLM-Importance & 0.608 & 0.020 & 0.625 & 0.032 & 0.657 & 0.023 \\
Current-task VoI (public) & 0.550 & 0.072 & 0.538 & 0.057 & 0.584 & 0.164 \\
\apa (ours) & 0.681 & 0.215 & 0.669 & 0.176 & 0.691 & 0.296 \\
Oracle Future-VoI & -- & -- & -- & -- & -- & -- \\
\bottomrule\end{tabular*}
\par\smallskip\begin{minipage}{\textwidth}\fontsize{7.4}{8.7}\selectfont
At $T=10$, NDCG is defined for 69 LLM-Importance users, 66 Current-task VoI (public) users and 68 APA users; Spearman is defined for all 71 users for these three policies. At $T=20$, both diagnostics use all 120 users for these policies.
\end{minipage}
\end{table*}

\section{Amortization-Enriched Test Set and APA-v2}
\label{app:amorttest}
Table~\ref{tab:amorthorizon} gives the horizon contrasts on the amortization-enriched test set (60 users, nested prefixes, $B=0.5$, $\lambda=0.5$, $\beta=0.25$, $K=10$), with Holm correction over its five horizons. The number of amortization-dependent pairs grows with the horizon (0, 19, 62, 202 and 210 at $T=1,3,5,10,20$); at $T=20$, \apa acquires 181 of the 210 pairs, Current-task VoI (public) none and the oracle 101. Because the original IID + unseen test set and the amortization-enriched test set contain different users, we compare their effects descriptively and do not test the difference between them.

\begin{table}[!htb]
\centering
\caption{Horizon contrasts on the amortization-enriched test set: \apa minus Current-task VoI (public), 60 paired users, pointwise 95\% CIs, Holm over five horizons.}
\label{tab:amorthorizon}
\fontsize{8.2}{9.5}\selectfont\setlength{\tabcolsep}{3.6pt}
\begin{tabular}{lrrr}
\toprule
$T$ & $\Delta$NU & 95\% CI & $p_{\rm Holm}$ \\
\midrule
1 & $-0.017$ & [$-0.040$, 0.002] & 0.364 \\
3 & $-0.015$ & [$-0.066$, 0.034] & 1.000 \\
5 & 0.030 & [$-0.057$, 0.129] & 1.000 \\
10 & 0.354 & [0.194, 0.533] & 0.0025 \\
20 & 1.296 & [0.786, 1.834] & 0.0025 \\
\bottomrule
\end{tabular}
\end{table}

\emph{APA-v2} adds a current-task value term to \apa's score: it scores a candidate by its current-task value, computed as in Current-task VoI (public), plus \apa's future value, minus $\lambda c_m+\beta C_{\text{risk}}(m)$; \emph{APA-v2 (double cost)} doubles both cost weights in the score only, while budgets and evaluation prices are unchanged. Table~\ref{tab:apav2} reports the paired NU difference to \apa, without significance tests. Neither variant improves NU consistently across horizons: \apa already acquires nearly all preferences that a single task justifies (241 of 242 at $T=20$ on the IID + unseen amortization cohort, and 89 of 89 at $T=10$ on the amortization-enriched set), so a current-task term adds little.

\begin{table}[!htb]
\centering
\caption{Mean paired NU difference of APA-v2 variants to \apa by horizon (180 IID + unseen or 60 amortization-enriched users; main settings). Means only.}
\label{tab:apav2}
\fontsize{8.2}{9.5}\selectfont\setlength{\tabcolsep}{2.4pt}
\begin{tabular*}{\columnwidth}{@{\extracolsep{\fill}}lrrrrr}
\toprule
Variant, test set & $T{=}1$ & 3 & 5 & 10 & 20 \\
\midrule
APA-v2, IID+unseen & 0.034 & 0.038 & 0.043 & $-0.007$ & $-0.013$ \\
APA-v2 (2$\times$cost), IID+unseen & 0.005 & 0.031 & 0.001 & $-0.003$ & 0.068 \\
APA-v2, amortization & 0.023 & 0.021 & 0.026 & 0.035 & $-0.013$ \\
APA-v2 (2$\times$cost), amortization & 0.009 & $-0.024$ & $-0.004$ & $-0.088$ & 0.102 \\
\bottomrule\end{tabular*}
\end{table}

\section{Generalization and Robustness}
\label{app:robust}
\subsection{IID / unseen-user results}
Tables~\ref{tab:iid} and~\ref{tab:unseen} separate the main comparison into its 90-user IID and unseen-user cohorts. Both use the default $B=0.5$ and report the same seven metrics as Table~\ref{tab:main}.
\begin{table*}[!tp]
\centering
\captionof{table}{IID: 90 users, $T=10$, $B=0.5$, $\lambda=0.5$, $\beta=0.25$, $K=10$. Definitions and regret references match Table~\ref{tab:main}. Cost-matched rows use the per-user caps of Table~\ref{tab:costmatched}; Random-$k$ averages five seeds.}
\label{tab:iid}
\fontsize{8}{9.4}\selectfont\setlength{\tabcolsep}{2.5pt}
\begin{tabular*}{\textwidth}{@{\extracolsep{\fill}}lrrrrrrr}
\toprule
Policy & CTU$\uparrow$ & AC$\downarrow$ & RC$\downarrow$ & NU$\uparrow$ & \#Q$\downarrow$ & Unn.$\downarrow$ & Regret$\downarrow$ \\
\midrule
Ask-Nothing & 6.430 & 0.000 & 0.000 & 6.430 & 0.000 & -- & 0.691 \\
Ask-Everything & 7.360 & 3.966 & 1.411 & 5.024 & 13.522 & 0.726 & 2.098 \\
LLM-Importance & 7.202 & 2.463 & 0.883 & 5.750 & 11.378 & 0.744 & 1.371 \\
Current-task VoI (public) & 6.959 & 0.383 & 0.000 & 6.768 & 2.067 & 0.332 & 0.353 \\
\apa (ours) & 7.191 & 0.694 & 0.056 & 6.830 & 3.800 & 0.387 & 0.291 \\
Oracle Future-VoI & 7.328 & 0.399 & 0.028 & 7.121 & 2.133 & 0.028 & 0.000 \\
\midrule
Random-$k$ (cost-matched) & 6.717 & 0.669 & 0.094 & 6.359 & 3.800 & 0.698 & 0.762 \\
Frequency (cost-matched) & 6.862 & 0.674 & 0.072 & 6.507 & 3.856 & 0.585 & 0.614 \\
Impact-only (cost-matched) & 7.153 & 0.683 & 0.083 & 6.791 & 3.867 & 0.472 & 0.330 \\
LLM-Importance (cost-matched) & 6.775 & 0.673 & 0.283 & 6.367 & 3.156 & 0.683 & 0.754 \\
\bottomrule\end{tabular*}
\end{table*}

\begin{table*}[!tp]
\centering
\captionof{table}{Unseen users: 90 users, $T=10$, $B=0.5$, $\lambda=0.5$, $\beta=0.25$, $K=10$. Definitions and regret references match Table~\ref{tab:main}. Cost-matched rows use the per-user caps of Table~\ref{tab:costmatched}; Random-$k$ averages five seeds. Cost-matched Unn. excludes one no-question user (89 defined users).}
\label{tab:unseen}
\fontsize{8}{9.4}\selectfont\setlength{\tabcolsep}{2.5pt}
\begin{tabular*}{\textwidth}{@{\extracolsep{\fill}}lrrrrrrr}
\toprule
Policy & CTU$\uparrow$ & AC$\downarrow$ & RC$\downarrow$ & NU$\uparrow$ & \#Q$\downarrow$ & Unn.$\downarrow$ & Regret$\downarrow$ \\
\midrule
Ask-Nothing & 6.149 & 0.000 & 0.000 & 6.149 & 0.000 & -- & 0.847 \\
Ask-Everything & 7.166 & 3.988 & 1.444 & 4.811 & 13.422 & 0.731 & 2.196 \\
LLM-Importance & 6.957 & 2.447 & 0.883 & 5.513 & 11.211 & 0.737 & 1.483 \\
Current-task VoI (public) & 6.747 & 0.376 & 0.000 & 6.560 & 2.022 & 0.295 & 0.436 \\
\apa (ours) & 7.027 & 0.665 & 0.061 & 6.680 & 3.633 & 0.387 & 0.316 \\
Oracle Future-VoI & 7.203 & 0.401 & 0.028 & 6.996 & 2.111 & 0.029 & 0.000 \\
\midrule
Random-$k$ (cost-matched) & 6.453 & 0.639 & 0.098 & 6.110 & 3.591 & 0.698 & 0.886 \\
Frequency (cost-matched) & 6.633 & 0.644 & 0.072 & 6.293 & 3.711 & 0.598 & 0.703 \\
Impact-only (cost-matched) & 6.920 & 0.649 & 0.083 & 6.574 & 3.622 & 0.452 & 0.422 \\
LLM-Importance (cost-matched) & 6.431 & 0.646 & 0.394 & 6.009 & 2.867 & 0.725 & 0.987 \\
\bottomrule\end{tabular*}
\end{table*}

\subsection{\texorpdfstring{Sensitivity to $\lambda$, $\beta$ and $B$}{Sensitivity to lambda, beta and B}}
\begin{figure*}[p]
\centering
\includegraphics[width=.84\textwidth]{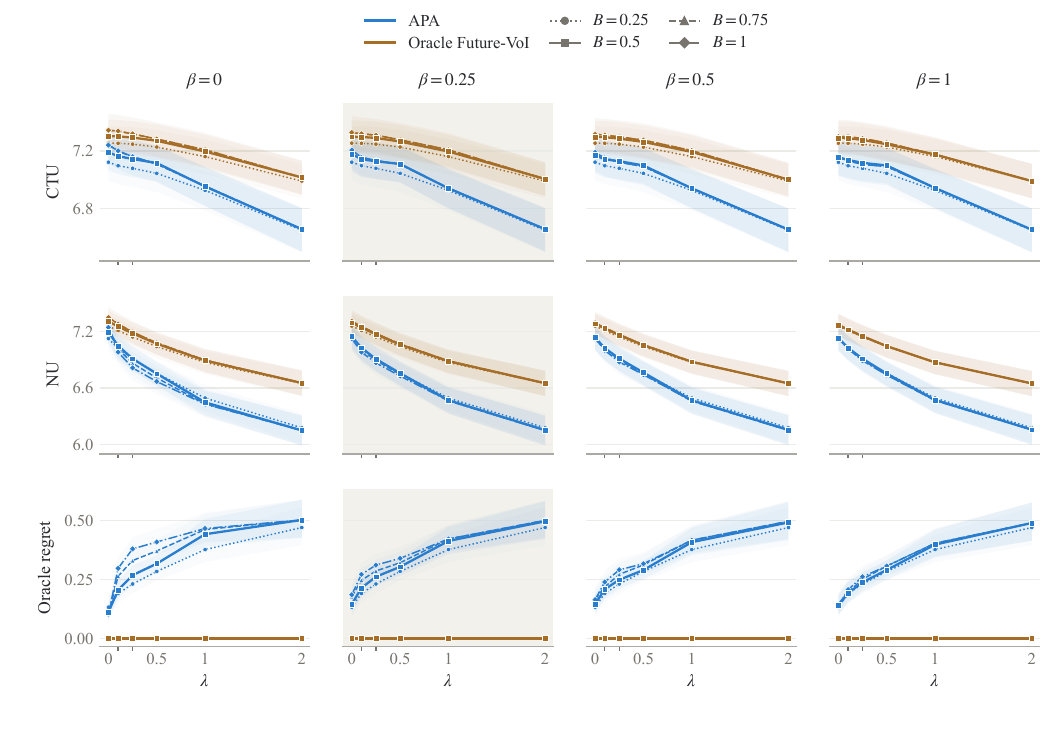}
\par\smallskip
\includegraphics[width=.84\textwidth]{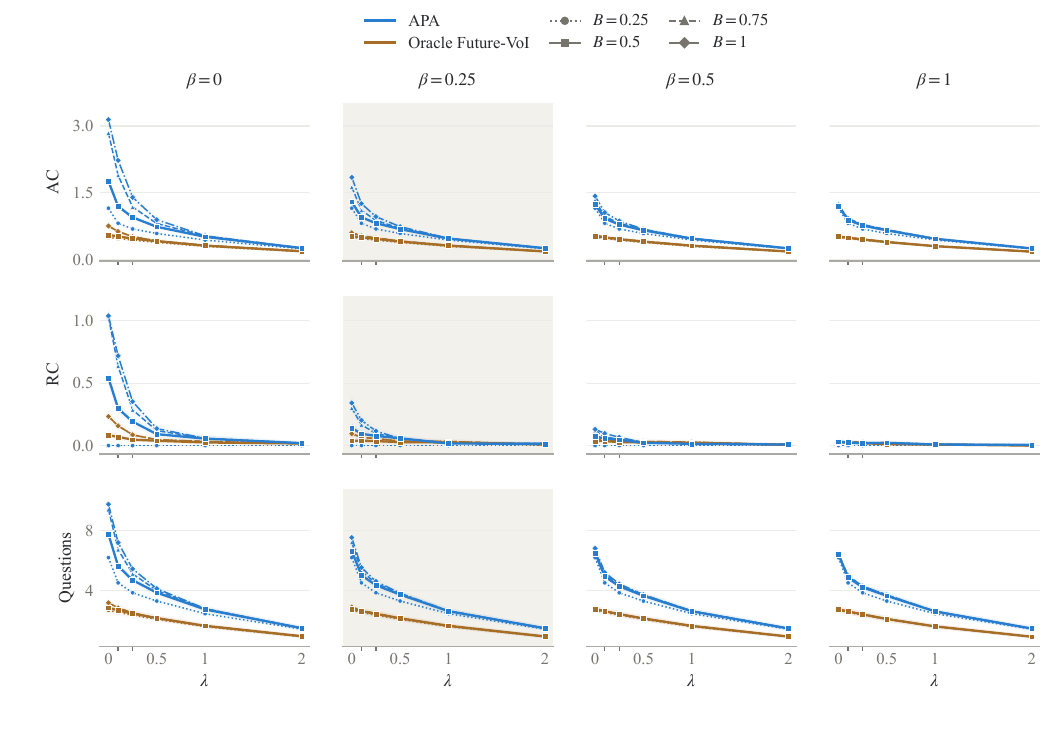}
\caption{Overall sensitivity (90 IID + 90 unseen users): performance (top: CTU, NU, regret) and behavior (bottom: AC, RC, questions). Each grid has $\beta\in\{0,0.25,0.5,1\}$ by column and $B\in\{0.25,0.5,0.75,1\}$ by line style; $\lambda\in\{0,0.1,0.25,0.5,1,2\}$ on each horizontal axis. $T=10$, $K=10$. Shading shows pointwise 95\% CIs. NU uses each cell\textquotesingle s own $\lambda$ and $\beta$. All saved means and CI endpoints are retained.}
\Description{Two compact Overall grids show all budget and coefficient combinations for APA and Oracle Future-VoI.}
\label{fig:sensitivity}
\end{figure*}

Figure~\ref{fig:sensitivity} includes the Overall performance and behavior grids for all 96 parameter settings per method. Columns vary $\beta$, line styles vary $B$, and the horizontal axis varies $\lambda$. Only the pooled 180-user results are displayed here.
\FloatBarrier

\FloatBarrier
\end{document}